\documentclass[a4paper,fleqn]{cas-sc}

\usepackage[authoryear,longnamesfirst]{natbib}

\usepackage{xcolor}

\usepackage{graphicx}
\usepackage{subcaption}
\usepackage{caption}

\usepackage{tabularx}
\usepackage{array}

\usepackage{booktabs} 

\usepackage{amssymb}

\usepackage{multirow}

\def\tsc#1{\csdef{#1}{\textsc{\lowercase{#1}}\xspace}}
\tsc{WGM}
\tsc{QE}
\tsc{EP}
\tsc{PMS}
\tsc{BEC}
\tsc{DE}

\begin{document}
\let\WriteBookmarks\relax
\def\floatpagepagefraction{1}
\def\textpagefraction{.001}

\shorttitle{Benchmarking Tesla’s Traffic Light and Stop Sign Control: Field Dataset and Behavior Insights}

\shortauthors{Zheng Li et~al.}

\title [mode = title]{Benchmarking Tesla’s Traffic Light and Stop Sign Control: Field Dataset and Behavior Insights}                      

\tnotetext[1]{This work was supported by the Federal Highway Administration (FHWA) under the Broad Agency Announcement (BAA) Award Number (Grant No. 693JJ324C000003). We gratefully acknowledge the support provided by the FHWA.}


%
\author[1,2]{Zheng Li}[
orcid=0000-0001-5732-5605,
]
\ead{zli2674@wisc.edu}

\author[1]{Peng Zhang}[]
\ead{pzhang257@wisc.edu}

\author[1]{Shixiao Liang}[]
\ead{sliang85@wisc.edu}

\author[1]{Hang Zhou}[]
\ead{hzhou364@wisc.edu}

\author[1]{Chengyuan Ma}[]
\ead{cma97@wisc.edu}

\author[3]{Handong Yao}[]
\ead{handong.yao@uga.edu}

\author[3]{Qianwen Li}[]
\ead{cami.li@uga.edu}

\author[1]{Xiaopeng Li}[]
\cormark[1]
\ead{xli2485@wisc.edu}



\affiliation[1]{organization={Department of Civil and Environmental Engineering, University of Wisconsin-Madison}, 
    city={Madison},
    postcode={53706}, 
    state={WI},
    country={United States}}

\affiliation[2]{organization={Department of Computer Science, University of Wisconsin-Madison}, 
    city={Madison},
    postcode={53706}, 
    state={WI},
    country={United States}}

\affiliation[3]{organization={School of Environmental, Civil, Agricultural and Mechanical Engineering, University of Georgia},
    city={Athens},
    postcode={30602}, 
    state={GA},
    country={United States}}

\cortext[cor1]{Corresponding author: Xiaopeng Li}



\begin{abstract}
Understanding how Advanced Driver-Assistance Systems (ADAS) interact with Traffic Control Devices (TCDs) is critical for assessing their influence on traffic operations, yet this interaction has received little focused empirical study. This paper presents a field dataset and behavioral analysis of Tesla’s Traffic Light and Stop Sign Control (TLSSC), a mature ADAS that perceives traffic lights and stop signs. We design and execute experiments across varied speed limits and TCDs types, collecting synchronized high-resolution vehicle trajectory data and driver-perspective video. From these data, we develop a taxonomy of TLSSC–TCD interaction behaviors (i.e., stopping, accelerating, and car following) and calibrate the Full Velocity Difference Model (FVDM) to quantitatively characterize each behavior mode. A novel empirical insight is the identification of a car-following threshold (\~90 m). Calibration results reveal that stopping behavior is driven by strong responsiveness to both desired speed deviation and relative speed, whereas accelerating behavior is more conservative. Intersection car-following behavior exhibits smoother dynamics and tighter headways compared to standard car-following behaviors. The established dataset, behavior definitions, and model characterizations together provide a foundation for future simulation, safety evaluation, and design of ADAS–TCD interaction logic. Our dataset is available at \href{https://github.com/CATS-Lab/Data-Tesla-Traffic-Light-and-Stop-Sign-Control}{\textcolor{blue}{Github}}.
\end{abstract}


\begin{highlights}
\item Developed a field dataset capturing Tesla’s Traffic Light and Stop Sign Control interactions with traffic lights and stop signs.  
\item Proposed a clear classification of system behaviors and calibrated a car-following model to characterize stopping, accelerating, and following.  
\item Discovered a car-following threshold of about 90 meters, revealing distinct behavioral dynamics at intersections.  
\end{highlights}

\begin{keywords}
Advanced Driver-Assistance Systems \sep traffic control devices \sep driving behavior modeling \sep autonomous driving dataset
\end{keywords}

\maketitle

\section{Introduction}

\par Autonomous Vehicles (AVs) have developed rapidly and are drawing increasing attention from both the public and researchers, driven by large-scale pilot programs and commercial deployments across major cities such as Phoenix, San Francisco, Los Angeles, and Austin. As AVs begin to operate in daily traffic, they introduce new challenges to the transportation system, including ensuring safety, managing Human–Machine Interaction (HMI), and adapting existing infrastructure and policies. To address these issues, a clear understanding of AV behavior and their impact on real-world environments is essential.

\par Existing autonomous driving technologies can generally be divided into two categories: higher-level Automated Driving Systems (ADS) and lower-level Advanced Driver-Assistance Systems (ADAS) (\cite{huang2023characterizing}). ADS aims for full automation within specific Operational Design Domains (ODD), for example, Tesla’s Full Self-Driving (FSD) and Lyft’s Level 5 ADS. In contrast, ADAS offers longitudinal and lateral driving support under constant human supervision. Common ADAS features include Adaptive Cruise Control (ACC), lane keeping, lane changing, detection of Traffic Control Devices (TCDs, such as traffic lights, stop signs, and speed limit signs), and basic interaction with these devices such as Tesla Autopilot’s Traffic Light and Stop Sign Control (TLSSC) feature.

\par The research community has extensively studied ADS behavior, including car following, lane changing, interactions with human drivers, and interactions with TCDs. For instance, \cite{hu2022processing} used the Waymo Open Dataset to extract car-following trajectories and analyze ADS car-following behaviors. \cite{ali2024investigating} investigated lane-changing characteristics using similar data sources. \cite{liu2024enhancing} applied a mixed-strategy game framework to model ADS turning behavior at intersections with interactions with human drivers. \cite{li2025interaction} examined ADS interactions with TCDs using Waymo Open Dataset and further analyzed these behaviors in their follow-up studies. Similarly, ADAS behavior has also been widely studied, primarily in the context of car following and lane changing. For example, \cite{huang2023characterizing} modeled ADAS car-following behaviors using the OpenACC dataset, while \cite{butakov2014personalized} analyzed lane-changing behaviors of ADAS-equipped vehicles. These studies also support system-level evaluations. For example, \cite{huang2023characterizing} assessed how large-scale deployment of ADS and ADAS technologies impacts traffic performance at the network level.

\par Despite extensive research on autonomous driving, little attention has been given to the interaction between ADAS and TCDs. Most ADAS-related studies have focused only on car-following and lane-changing behaviors, while investigations of AV–TCD interactions have mainly centered on ADS systems. This oversight may be due to two reasons. First, ADAS systems capable of interacting with TCDs have been relatively rare in the past years, as many ADAS-equipped vehicles lacked the ability to detect or respond to such devices earlier. Second, the absence of publicly available datasets capturing ADAS–TCD interactions has made it difficult to conduct systematic research in this area.

\par Yet, as commercial vehicles continue to evolve, several production models now offer traffic light and traffic sign detection combined with longitudinal control, and these systems are already operating on public roads. For example, Volvo Trucks are equipped with Volvo Active Driver Assist (VADA), which includes a “Road Sign Recognition” feature that uses forward-facing cameras to identify various traffic signs, such as speed limits, and display them on the dashboard (\cite{Volvo_Trucks}). Similarly, Freightliner Cascadia trucks equipped with the Detroit Assurance suite (ABA6/ABA5) include a “Traffic Sign Display” function that reads and displays signs like speed limits in real time, while combining high-definition cameras and radar to detect stationary vehicles and deceleration scenarios (\cite{Freightliner_Cascadia}). Ford Pro’s commercial vehicles (e.g., Transit and F-Series used in fleet settings) feature Co-Pilot360 with Speed Sign Recognition, which identifies speed signs and integrates with cruise control (\cite{Ford_Pro}). Among these, Tesla’s Autopilot system with Traffic Light and Stop Sign Control (TLSSC) stands out for its advanced capabilities, not only detecting TCDs with high accuracy, but also executing appropriate interactions. TLSSC uses forward-facing cameras and sensors to identify traffic lights and stop signs, automatically decelerating, stopping, and resuming motion when safe and permitted (\cite{tesla_tlss}).

\par It could be found that many commercial vehicles equipped with ADAS now have the ability to detect and interact with TCDs. However, little research has been conducted to analyze these interactions, and publicly available data supporting such studies is lacking so far. To address this gap, this study focuses on Tesla’s TLSSC system, one of the most advanced ADAS platforms for TCD detection and interaction. 

\par We design and conduct extensive field experiments to collect detailed interaction data and build a dedicated dataset. Using this dataset, we systematically analyze the behavior of TLSSC-enabled Vehicles (TLSSC-V) during interactions with traffic lights and stop signs. We conduct experiments at a variety of intersections with different speed limits to capture a wide range of interaction scenarios. The interaction behaviors we focused includes stopping behaviors, accelerating behaviors, and car-following behaviors. The TLSSC-V we utilized to observe its behavior is equipped with high-frequency GPS and a forward-facing camera to collect trajectory and visual data. We synchronize video and GPS trajectory data, apply smoothing and quality assessments, and annotate key events such as stopping moments, signal changes, and permission cues. Finally, we fit the Full Velocity Difference Model (FVDM) to characterize behavioral dynamics.

\par This work makes three key contributions. First, we develop a field dataset that captures ADAS interactions with TCDs using Tesla’s TLSSC system, featuring labeled trajectory segments synchronized with driver-perspective video across diverse intersection types, speed limits, and behavioral scenarios. Second, we propose a behavior taxonomy along with precise operational definitions to categorize TLSSC responses to traffic lights and stop signs. Third, we adopt a model-based approach to characterize TLSSC behavior by calibrating the FVDM for various behavior types and analyzing behavioral characteristics based on the calibration results.

\par The remainder of this paper is organized as follows. Section 2 reviews related datasets relevant to our study. Section 3 provides a detailed description of the TLSSC behaviors under investigation. Section 4 outlines our experimental design for capturing these behaviors. Section 5 presents the data processing pipeline. Section 6 summarizes the TLSSC dataset we constructed. Then, Section 7 analyzes the observed TLSSC behaviors. Finally, Section 8 concludes this work.

\section{Relevant dataset review}
\par To demonstrate the necessity of creating a new dataset for studying ADAS–TCD interaction behaviors, we review existing related datasets.

\par First, we surveyed publicly available ADS datasets that include interactions with TCDs. The datasets we focused on include: nuScenes, nuPlan, KITTI, and Waymo Open.

\par nuScenes dataset (\cite{caesar2020nuscenes}): This dataset was collected using two Renault Zoe electric vehicles operating in urban areas of Boston and Singapore. It includes LiDAR and camera data capturing AV trajectories, surrounding vehicle movements, and environmental context, with a focus on urban intersections. While the dataset contains numerous scenarios involving interactions with traffic lights and stop signs, all recorded behaviors are associated with ADS, and no ADAS-equipped vehicles were observed.

\par nuPlan (\cite{caesar2021nuplan}): Similar in scope to nuScenes, nuPlan extends its capability by developing a novel method to infer traffic light states from vehicle motion patterns. Although this dataset includes trajectories from both ADS and Human-driven Vehicles (HVs), it does not contain identifiable instances of ADAS-equipped vehicle behavior.

\par KITTI (\cite{geiger2013vision}): KITTI is primarily designed for AV perception research. The data was collected using a Volkswagen station wagon and includes approximately six hours of multi-sensor traffic data. The scenarios span various road types, from highways and rural roads to urban streets, with rich interactions involving static and dynamic objects. Despite containing many segments where vehicles interact with TCDs, the recorded behaviors are from ADS platforms, not ADAS.

\par Waymo Open (\cite{sun2020scalability}): This dataset provides high-resolution sensor data collected by Waymo vehicles operating in cities such as Phoenix and Mountain View. It covers a wide range of traffic scenarios, including car following, lane changing, gap acceptance, pedestrian interaction, and interactions with traffic lights and signs. However, similar to the above datasets, the recorded vehicles are ADS-equipped, and there is no evidence of ADAS systems being involved.

\par We also reviewed several mainstream ADAS-related datasets, including OpenACC and Ohio.

\par OpenACC (\cite{makridis2021openacc}): OpenACC dataset was developed to investigate commercial ACC systems. The data were collected from both public highways and controlled test tracks across various locations in Europe, capturing detailed vehicle trajectory information such as speed, acceleration, and precise positioning using high-accuracy sensors and onboard systems. However, this dataset primarily focuses on the longitudinal control functionality of ADAS and provides extensive platoon-level longitudinal trajectory data. It does not include any scenarios or data involving ADAS interactions with TCDs.

\par Ohio (\cite{seitz2024advanced}): The Ohio dataset provides naturalistic driving data collected from passenger vehicles equipped with ADAS technologies. It includes detailed vehicle trajectories, GPS traces, and control information across various road types and traffic conditions in Ohio. While the dataset contains scenarios involving TCDs, most interactions with TCDs were performed by human drivers, as the vehicles in the dataset did not consistently possess the capability to detect or respond to TCDs. As a result, like the OpenACC dataset, the Ohio dataset primarily emphasizes longitudinal control and general driving behaviors rather than focused analysis of ADAS–TCD interactions.

\par Based on the above analysis, it is evident that there is a lack of datasets capturing direct interactions between ADAS and TCDs. Existing datasets on AV–TCD interactions exclusively involve ADS-equipped vehicles, while datasets focused on ADAS behavior largely omit scenarios involving direct engagement with TCDs. This highlights the necessity of developing a dedicated dataset that captures ADAS–TCD interactions.

\section{Behavior clarification}
\par As previously mentioned, we use Tesla’s TLSSC-V to collect behavioral data on ADAS–TCD interactions and analyze its responses. In this section, we describe the behavioral characteristics of TLSSC-V. All behavior descriptions are based on Tesla’s official documentation for the TLSSC feature (\cite{tesla_tlss}) and our own observations during the data collection process. Specifically, we categorize and clarify three types of TLSSC-V behaviors when encountering traffic lights and stop signs: stopping behavior, accelerating behavior and car-following behavior.

\subsection{Stopping behavior}
\par The following summarizes the stopping behaviors exhibited by the TLSSC-V when encountering traffic lights and stop signs. We identify three distinct types: stop before a red and yellow light, stop before a green light, and stop before a stop sign.

\begin{itemize}
    \item \textbf{Stop before a red and yellow light}: refers to the situation where, upon detecting a red or yellow traffic light ahead, the TLSSC-V displays a red stop line on the touchscreen, then slows down and comes to a complete stop at the indicated line.

    \item \textbf{Stop before a green light}: refers to the situation where, when TLSSC-V detects a green traffic light ahead and there is no lead vehicle in front, it slows down and waits for the human driver to grant permission to proceed through the intersection. If the driver does not grant permission, TLSSC-V stops at the red stop line shown on the touchscreen, even though the light is green.

    \item \textbf{Stop before a stop sign}: refers to the situation where, when TLSSC-V detects a stop sign ahead, it slows down and comes to a complete stop at the red stop line displayed on the touchscreen.
\end{itemize}

\subsection{Accelerating behavior}
\par We categorize the observed acceleration behaviors into two types: accelerate after permission at a green light and accelerate after permission at a stop sign.

\begin{itemize}
    \item \textbf{Accelerate after permission at a green light}: refers to cases where the TLSSC-V slows down after detecting an ahead green light, receives permission from the human driver to proceed. The vehicle then accelerates and resumes its desired speed to pass through the intersection. Alternatively, the TLSSC-V may continue slowing down when a green light is detected until coming to a complete stop at the stop line; once the human driver grants permission, the vehicle accelerates and proceeds through the intersection. Therefore, the accelerating behavior here can be further divided into two types: \textbf{accelerate before a stop} and \textbf{accelerate after a stop}.

    \item \textbf{Accelerate after permission at a stop sign}: refers to the process in which the TLSSC-V comes to a complete stop at the stop line after detecting a stop sign, then accelerates through the intersection once the human driver grants permission.
\end{itemize}

\subsection{Car-following behavior}
\par We classify the observed car-following behaviors of the TLSSC-V into two categories: standard car-following behavior and car-following behavior when proceeding straight through the intersection.

\begin{itemize}
    \item \textbf{Standard car-following behavior}: refers to the car-following behavior exhibited by the TLSSC-V when it is not interacting with a traffic light or stop sign, i.e., during segments of the road that are relatively far from the intersection.
    
    \item \textbf{Car-following behavior when proceeding straight through the intersection}: refers to the situation where the TLSSC-V intends to go straight through an intersection during a green light and follows a lead vehicle along its path as it passes through the intersection.
\end{itemize}

\subsection{Human intervention}
\par In the previous behavior descriptions, we frequently mentioned the process of the human driver granting permission. This is because TLSSC is fundamentally an ADAS feature, a driver-assistance system, not an ADS or FSD. As such, interactions between TLSSC and traffic lights or stop signs may require human driver intervention. We present two forms of human driver involvement in the interactions with TLSSC (\cite{tesla_tlss}):

\begin{itemize}
    \item \textbf{Grant permission}: this refers to situations where the TLSSC-V approaches an intersection with a green light but, in the absence of a lead vehicle, slows down and waits for the human driver to lightly press the accelerator pedal to authorize proceeding. Similar behavior occurs when TLSSC-V stops at a red light; once the light turns green and the TLSSC-V intends to proceed straight through the intersection, it waits for the driver’s input to continue. Likewise, at a stop sign, when the TLSSC-V intends to go straight through the intersection, it comes to a complete stop and only proceeds after the human driver briefly presses the accelerator pedal to grant permission.
    
    \item \textbf{Take control}: This occurs when an TLSSC-V vehicle is unable to independently complete the decision-making and maneuvering required for a turn, prompting the human driver to take control the vehicle and to drive.
\end{itemize}

\par It is important to note that this study focuses specifically on ADAS behavior. Therefore, we only examine the behavior of the TLSSC system itself. This includes actions taken by TLSSC both before and after the human driver grants permission. However, we do not analyze any behavior that occurs after the human driver takes full control of the vehicle.

\section{Experiment}
\par This section provides a detailed explanation of the experimental setup we designed to collect the TLSSC behavioral data.

\par
\begin{figure}[htbp]
    \centering
    \begin{subfigure}[t]{0.32\textwidth}
        \centering
        \includegraphics[height=2.8cm]{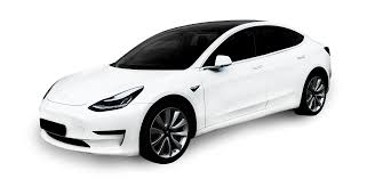}
        \caption{TLSSC-V.}
        \label{TLSSC-V}
    \end{subfigure}
    \hfill
    \begin{subfigure}[t]{0.32\textwidth}
        \centering
        \includegraphics[height=2.8cm]{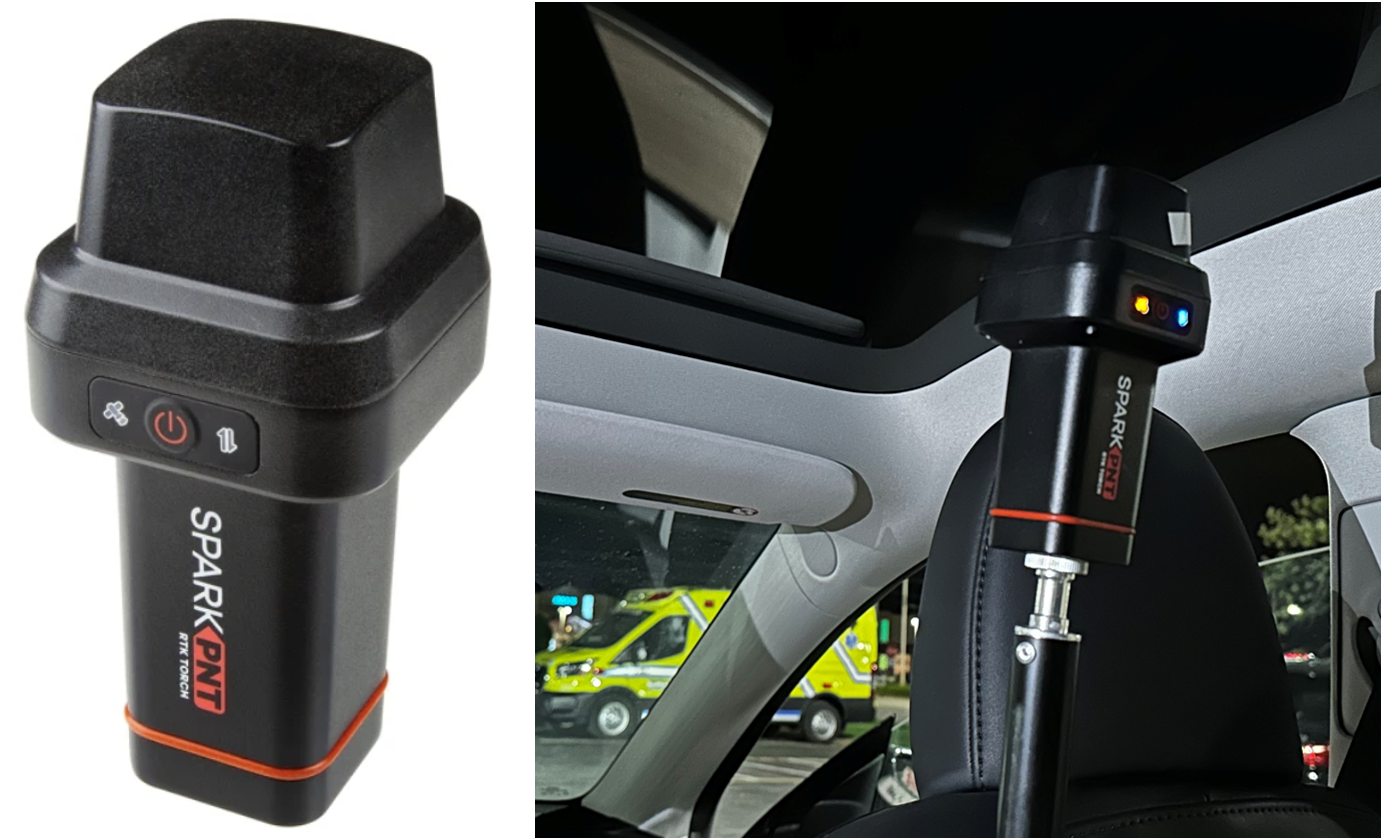}
        \caption{SparkFun Torch GPS.}
        \label{SparkFunGPS}
    \end{subfigure}
    \hfill
    \begin{subfigure}[t]{0.32\textwidth}
        \centering
        \includegraphics[height=2.8cm]{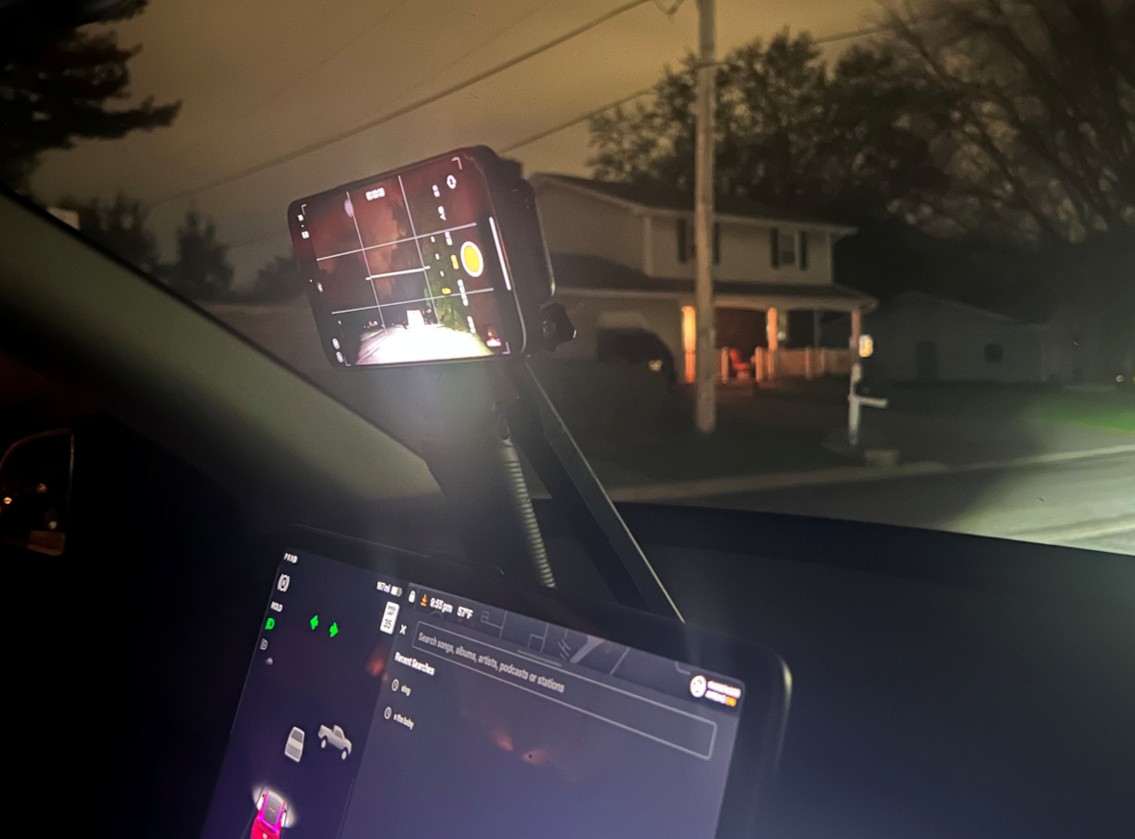}
        \caption{In-vehicle camera.}
        \label{InVehicleCamera}
    \end{subfigure}
    \caption{Equipments to collect the behavior data for TLSSC.}
    \label{equipment}
\end{figure}

 Equipments to collect the behavior data for TLSSC is shown in Figure \ref{equipment}. We used a Tesla Model Y equipped with the TLSSC feature (referred to as TLSSC-V throughout this paper) to collect data and observe behaviors. Multiple SparkFun Torch GPS devices were used to record the vehicle’s trajectory data, with a temporal resolution of 0.1 seconds. Additionally, a camera was mounted on the front windshield of the TLSSC-V to capture driver-perspective video throughout the data collection process. 

\par All data were collected at various intersections across four cities in Wisconsin, U.S.: Madison, Middleton, Fitchburg, and Verona. The selected intersections were diverse, including signalized intersections and stop sign-controlled intersections, with connecting roads having speed limits ranging from 25 to 50 mph. Since TLSSC can only be activated on structured roads with clearly visible signs and lane markings, all data were collected on well-marked, structured roadways. To ensure safety and minimize disruption to surrounding traffic, all experiments were conducted at night.

\par The specific setups for observing each TLSSC behavior are described as follows.

\subsection{Setup for stopping behavior observation}

\par For the stop before a red and yellow light and stop before a green light behaviors, we selected three intersections with approach lane speed limits of 25, 35, and 40 mph, respectively. The desired speed of the TLSSC-V was set to match the posted speed limit. The TLSSC-V repeatedly passed through each intersection multiple times, allowing us to record its stopping behavior at traffic lights under different speed limits.

\par For the stop before a stop sign behavior, we selected one intersection with an approach lane speed limit of 50 mph. We configured the TLSSC-V with desired speeds of 50, 45, 35, and 25 mph, and had the vehicle repeatedly pass through the intersection. This setup allowed us to observe and record the stopping behavior in front of the stop sign under varying desired speeds.

\subsection{Setup for accelerating behavior observation}

\par For the accelerate after permission at a green light behavior, we selected three intersections with approach lane speed limits of 25, 35, and 40 mph, respectively. The desired speed of the TLSSC-V was set to match the posted speed limit. In this observation, we designed two experimental groups. In the first group, TLSSC-V was brought to a complete stop at the stop line during a green light, after which the human driver granted permission to proceed allowing us to observe acceleration behavior from a standstill. In the second group, the human driver granted permission while the TLSSC-V was decelerating toward the green light but had not yet come to a full stop, enabling observation of its acceleration behavior from a rolling state. The TLSSC-V repeatedly passed through each intersection multiple times, allowing us to record its acceleration behavior at traffic lights under varying speed limits.

\par For the accelerate after permission at a stop sign behavior, we selected one intersection with an approach lane speed limit of 35 mph. The TLSSC-V was configured with desired speeds of 35 and 25 mph. In each trial, the TLSSC-V was brought to a complete stop at the stop line, after which the human driver granted permission for the vehicle to accelerate from a standstill and proceed through the intersection. The TLSSC-V repeatedly passed through the intersection multiple times, allowing us to record its acceleration behavior under different desired speeds.

\subsection{Setup for car-following behavior observation}

\par
\begin{figure}[h!t]
\centering
\includegraphics[width=0.6\linewidth]{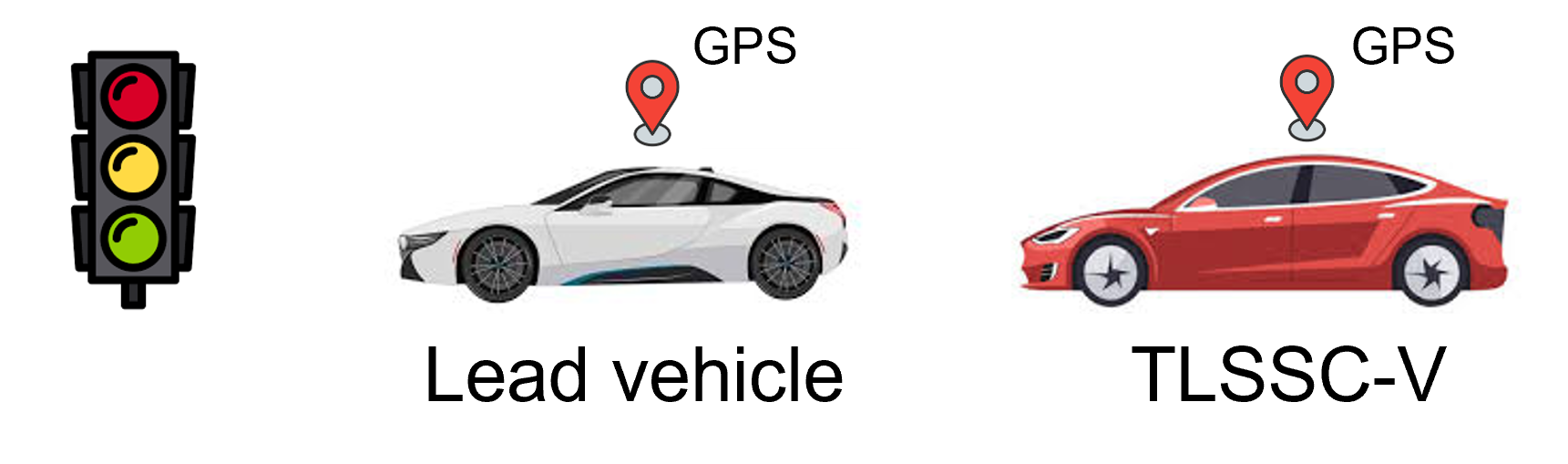}
\caption{Setup for the TLSSC-V car-following behavior observation.}
\label{Car_Following_Setting}
\end{figure}

 To observe the car-following behavior of the TLSSC-V, we introduced a 2020 Subaru Outback Limited as the lead vehicle. The TLSSC-V followed the lead vehicle during the trials, as illustrated in Figure \ref{Car_Following_Setting}.

\par To examine the characteristics of the standard car-following behavior, we designed an oscillatory scenario in which the lead vehicle’s speed varied sequentially through the following pattern: 40 mph → 30 mph → 20 mph → 30 mph → 40 mph. This setup allowed us to observe the TLSSC-V’s car-following responses under dynamic speed changes. Notably, the TLSSC-V used in our experiments supports discrete gap level settings ranging from 2 to 7, which control the following distance. In this experiment, we tested gap levels of 2, 4, and 7 to analyze car-following behavior under different following distances.

\par For the car-following behavior when proceeding straight through the intersection, we selected a series of consecutive intersections, each with an approach lane speed limit of 40 mph. The TLSSC-V’s desired speed was set to 40, 30, 25, and 20 mph across different trials, and the gap level was varied among 7, 4, and 2. The TLSSC-V followed the lead vehicle through these intersections multiple times, enabling us to observe its car-following behavior under different speed and gap level settings when passing through intersections.

\section{Data processing}

\par We conducted a series of post-processing steps on the collected data to ensure its accuracy and usability for behavior analysis. The data processing details are as follows.

\subsection{Data preprocessing}

\par First, we extracted valid trajectory segments from the raw GPS data, filtering out irrelevant or noisy portions of the data. Next, we interpolated missing values in the GPS data to ensure temporal continuity, followed by trajectory smoothing using a moving-average filter to reduce measurement noise and enhance the clarity of motion patterns. We also evaluated the quality of the processed trajectories to ensure they were reliable and suitable for studying TLSSC behaviors. Finally, the GPS trajectories were temporally synchronized with the in-vehicle camera video, enabling frame-level alignment between vehicle movement and visual context.

\par Besides, we manually annotated several key timestamps, including stop time, green light time, human permission time, and the position of the stop line in each valid trajectory segment. These annotations provide critical reference points for analyzing the TLSSC-V’s behavior. Specifically, they allow us to quantify the vehicle's reaction time after a red light turns green, as well as the delay between the human driver’s input and the vehicle’s response during permission-based interactions. This information is essential for understanding both the system's responsiveness and the interaction between TLSSC and human driver.

\subsection{Data enhancement}

\par To reduce noise and improve the clarity of behavioral patterns in the raw GPS trajectories, we applied a moving average filter with a window size of 1 second. Given that the GPS sampling rate is 10 Hz (i.e., one data point every 0.1 seconds), the smoothing window includes 10 consecutive points.

\par The smoothed value at time step \( t \), denoted as \( \bar{z}_t \), is computed as the average of the original values over a symmetric window centered at \( t \). The formula of the moving average filter is given by Equation \ref{eq:moving_average}. We applied smoothing to the latitude, longitude, and speed in the GPS trajectory, so \(z\) in this Equation refers to the features of latitude, longitude, and speed. 

\begin{equation}
\bar{z}_t = \frac{1}{N} \sum_{i = t - \lfloor N/2 \rfloor}^{t + \lfloor N/2 \rfloor} z_i
\label{eq:moving_average}
\end{equation}
\medskip

\par where \( z_i \) is the original feature value at time step \( i \), and \( N = 10 \) is the number of samples in the 1-second window.

\subsection{Data assessment}

\par To ensure that the trajectory data contained minimal noise and outliers and accurately reflected the characteristics of the TLSSC-V, we conducted a quality assessment of the trajectory data.

\par Following our previous related work, we used two metrics to evaluate the quality of the trajectory data  (\cite{li2025interaction}): Anomaly Acceleration (\%), and Anomaly Jerk (\%), which provide a quantitative assessment of deviations from expected behavior in terms of acceleration and jerk (\cite{li2023large}).

\par Anomaly Acceleration (\%) measures the percentage of acceleration \(a\) values that are considered anomalous. An acceleration value is deemed anomalous if it exceeds a predefined threshold. The normal range of acceleration is \(a \in [-8ms^{-2}, 5ms^{-2}] \) (\cite{punzo2011assessment}).

\par Anomaly Jerk (\%) quantifies the percentage of jerk data that are identified as anomalous. Jerk (\(j\)) is the rate of change in acceleration over time, as shown in Equation \ref{jerk definition}. Similar to acceleration, a jerk value is considered anomalous if it exceeds a specified threshold. The normal range of acceleration is \(j \in [-15ms^{-3}, 15ms^{-3}] \) (\cite{punzo2011assessment}).

\begin{equation}
    \label{jerk definition}
    j = \frac{da}{dt}
\end{equation}
\medskip  

\section{Dataset introduction}
\par Following the data collection procedures described previously, we collected a large amount of TLSSC behavior data. After processing and enhancement, we constructed the TLSSC dataset. This section introduces the TLSSC dataset including a summary, data format, and the quality of the trajectory data.

\subsection{Dataset summary}

\par Table \ref{tab:quantity summary} presents a summary of the quantity of trajectory segments collected for each TLSSC behavior under different speed settings in the TLSSC dataset. The rows represent various types of TLSSC behaviors, including stopping and accelerating behaviors as well as car-following behaviors with different gap levels. The columns correspond to the desired speed settings (ranging from 20 mph to 50 mph) under which the data were collected. The Table \ref{tab:quantity summary} provides an overview of the coverage and distribution of the TLSSC dataset across different behavioral scenarios and desired speed conditions.

\begin{table}[h]
    \centering
    \caption{Trajectory segment quantity for each TLSSC behavior in TLSSC dataset.}
    \begin{tabular}{|>{\centering\arraybackslash}m{5.5cm}|>{\centering\arraybackslash}m{1cm}|>{\centering\arraybackslash}m{1cm}|>{\centering\arraybackslash}m{1cm}|>{\centering\arraybackslash}m{1cm}|>{\centering\arraybackslash}m{1cm}|>{\centering\arraybackslash}m{1cm}|>{\centering\arraybackslash}m{1cm}|}
    \hline
         TLSSC behavior& 20 mph& 25 mph& 30 mph& 35 mph& 40 mph& 45 mph& 50 mph\\ \hline
         Stop before a red and yellow light&  &  3&  &  3&  7&   & \\ \hline
         Stop before a green light&  &  3&  &  3&  3&   & \\ \hline
         Stop before a stop sign&  &  3&  &  3&  &   3&3 \\ \hline
         Accelerate after permission at a green light (Before a stop)&  &  4&  &  3&  3&   & \\ \hline
         Accelerate after permission at a green light (After a stop)& & 3& & 3& 3&  & \\ \hline
         Accelerate after permission at a stop sign&  1&  &  1&  &  2&   & \\ \hline
         Standard car-following behavior (Gap level 2) & \multicolumn{7}{c|}{1} \\ \hline
         Standard car-following behavior (Gap level 4) & \multicolumn{7}{c|}{1} \\ \hline
         Standard car-following behavior (Gap level 7) & \multicolumn{7}{c|}{1} \\ \hline
 Car-following behavior when proceeding straight through the intersection (gap level 2)& 3& & 3& & 3&  & \\ \hline
 Car-following behavior when proceeding straight through the intersection (gap level 4)& 3& & 3& & 4& &\\ \hline
 Car-following behavior when proceeding straight through the intersection (gap level 7)& & 3& 3& & 3& &\\ \hline
    \end{tabular}
    \label{tab:quantity summary}
\end{table}

\par
\begin{figure}[htbp]
    \centering
    
    \begin{subfigure}[t]{0.18\textwidth}
        \centering
        \includegraphics[height=7cm]{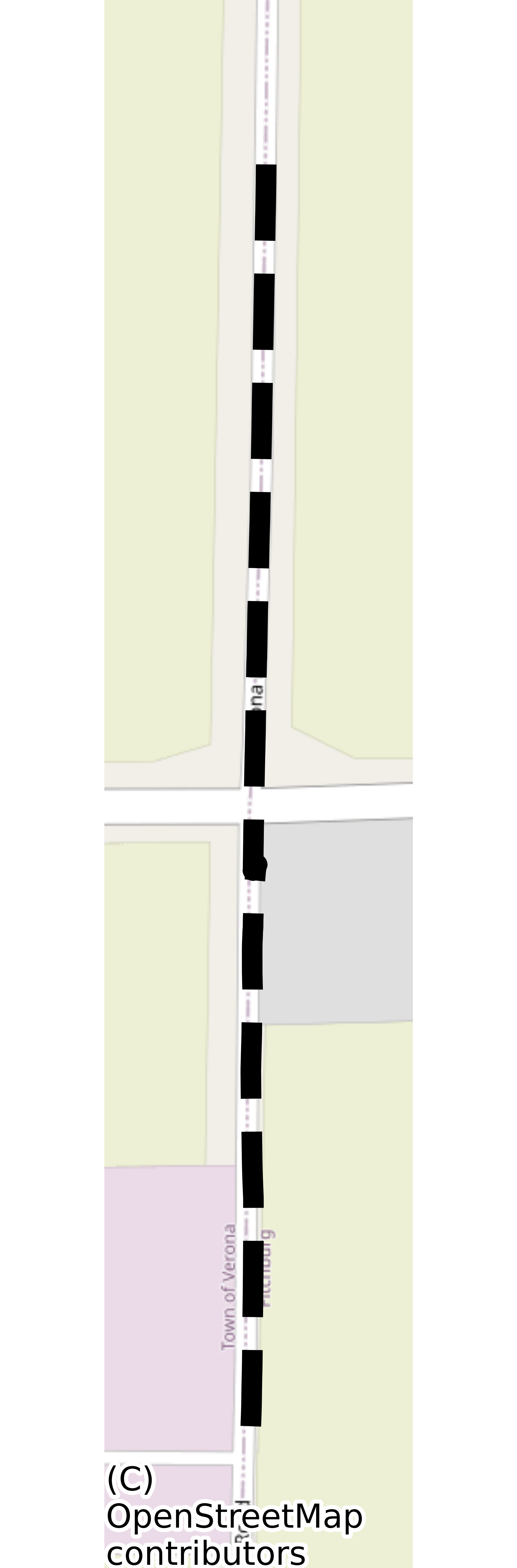}
        \caption{Stop sign interaction behavior - Example 1}
    \end{subfigure}
    \begin{subfigure}[t]{0.18\textwidth}
        \centering
        \includegraphics[height=7cm]{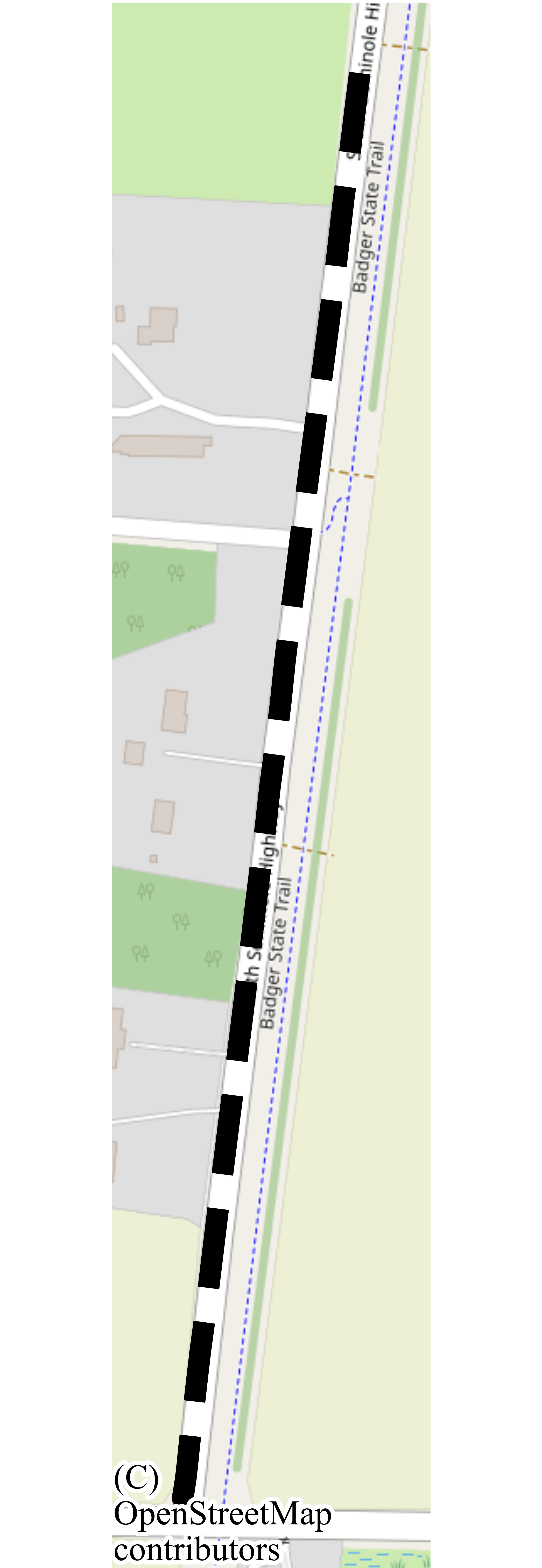}
        \caption{Stop sign interaction behavior - Example 2}
    \end{subfigure}
    \begin{subfigure}[t]{0.18\textwidth}
        \centering
        \includegraphics[height=7cm]{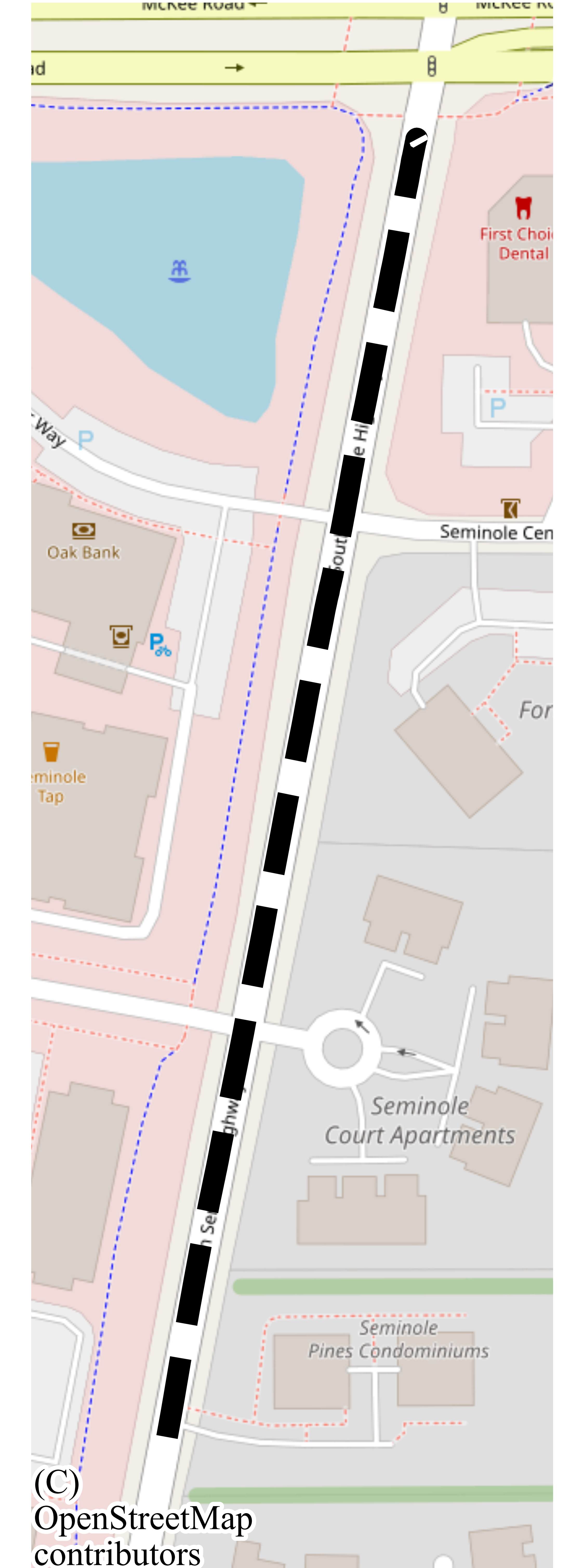}
        \caption{Traffic light interaction behavior - Example 1}
    \end{subfigure}
    \begin{subfigure}[t]{0.18\textwidth}
        \centering
        \includegraphics[height=7cm]{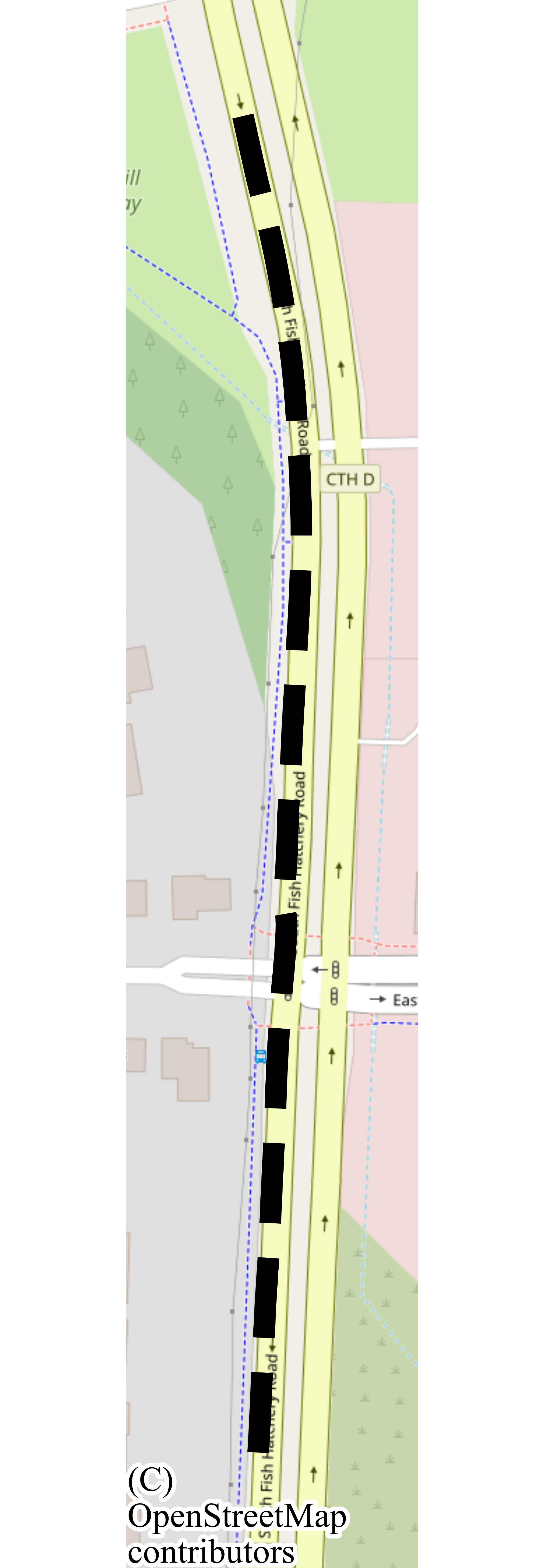}
        \caption{Traffic light interaction behavior - Example 2}
    \end{subfigure}

    \caption{Examples of position profiles in the TLSSC dataset.}
    \label{fig:position profile}
\end{figure}

 Figure \ref{fig:position profile} shows examples of position profiles from the TLSSC dataset we constructed. Since some data may have been collected at the same intersection, plotting all position profiles could result in redundancy. Therefore, we selected only four representative examples: two illustrate TLSSC-V’s behavior at traffic lights, and the other two depict its behavior at stop signs.

\par
\begin{figure}[htbp]
    \centering

    \begin{subfigure}[t]{0.3\textwidth}
        \centering
        \includegraphics[height=4cm]{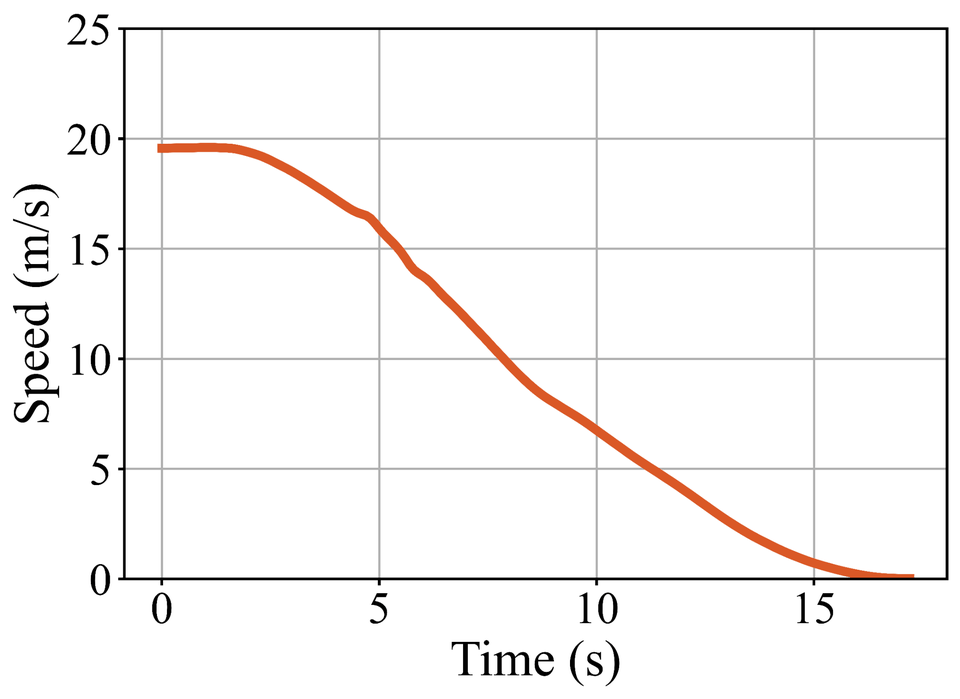}
        \caption{Stop before a red and yellow light.}
    \end{subfigure}
    \hfill
    \begin{subfigure}[t]{0.3\textwidth}
        \centering
        \includegraphics[height=4cm]{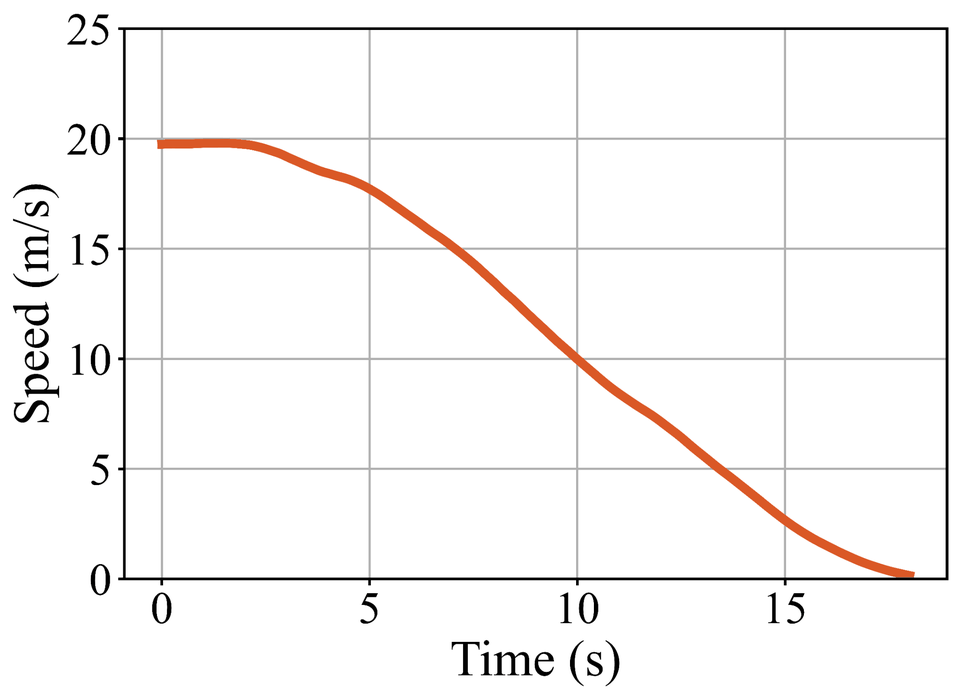}
        \caption{Stop before a green light.}
    \end{subfigure}
    \hfill
    \begin{subfigure}[t]{0.3\textwidth}
        \centering
        \includegraphics[height=4cm]{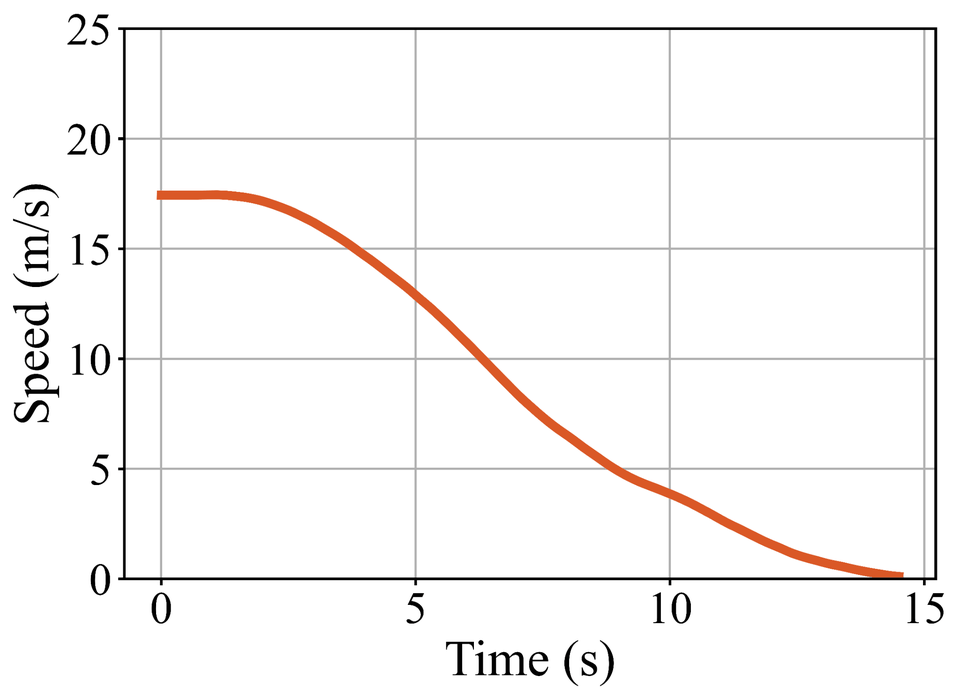}
        \caption{Stop before a stop sign.}
    \end{subfigure}

    \vspace{1em} 
    
    \begin{subfigure}[t]{0.3\textwidth}
        \centering
        \includegraphics[height=4cm]{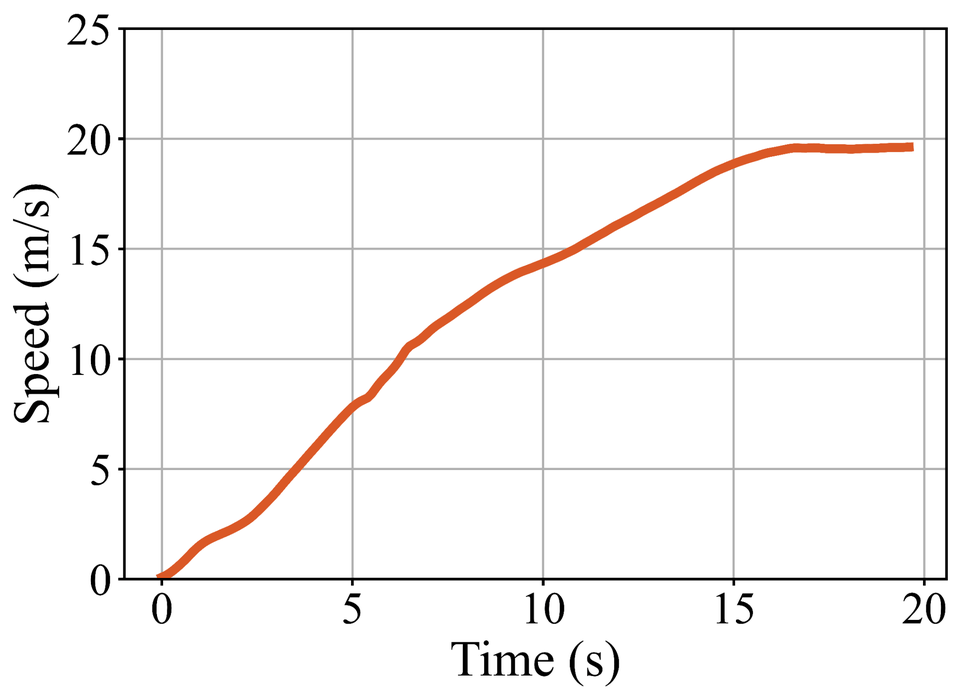}
        \caption{Accelerate after permission at a green light (After a stop).}
    \end{subfigure}
    \hfill
    \begin{subfigure}[t]{0.3\textwidth}
        \centering
        \includegraphics[height=4cm]{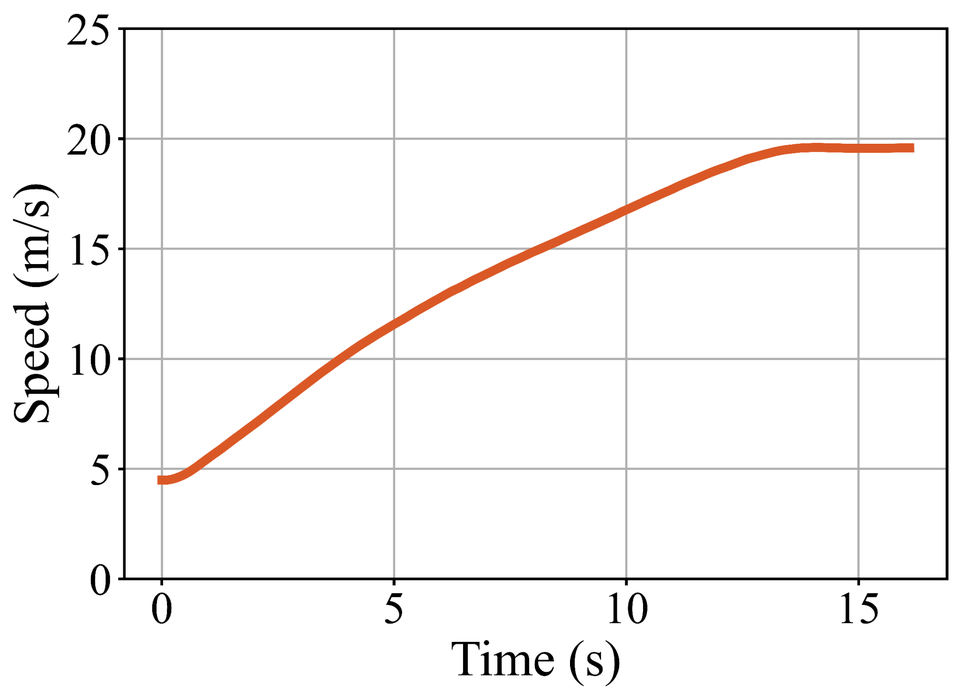}
        \caption{Accelerate after permission at a green light (Before a stop).}
    \end{subfigure}
    \hfill
    \begin{subfigure}[t]{0.3\textwidth}
        \centering
        \includegraphics[height=4cm]{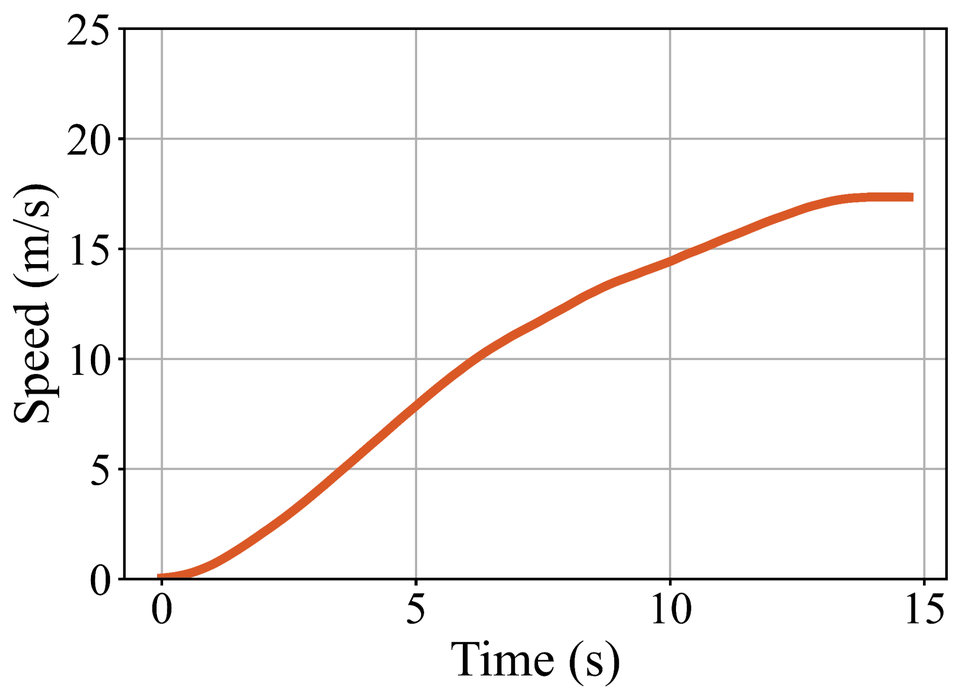}
        \caption{Accelerate after permission at a stop sign.}
    \end{subfigure}

    \begin{subfigure}[t]{0.3\textwidth}
        \centering
        \includegraphics[height=4cm]{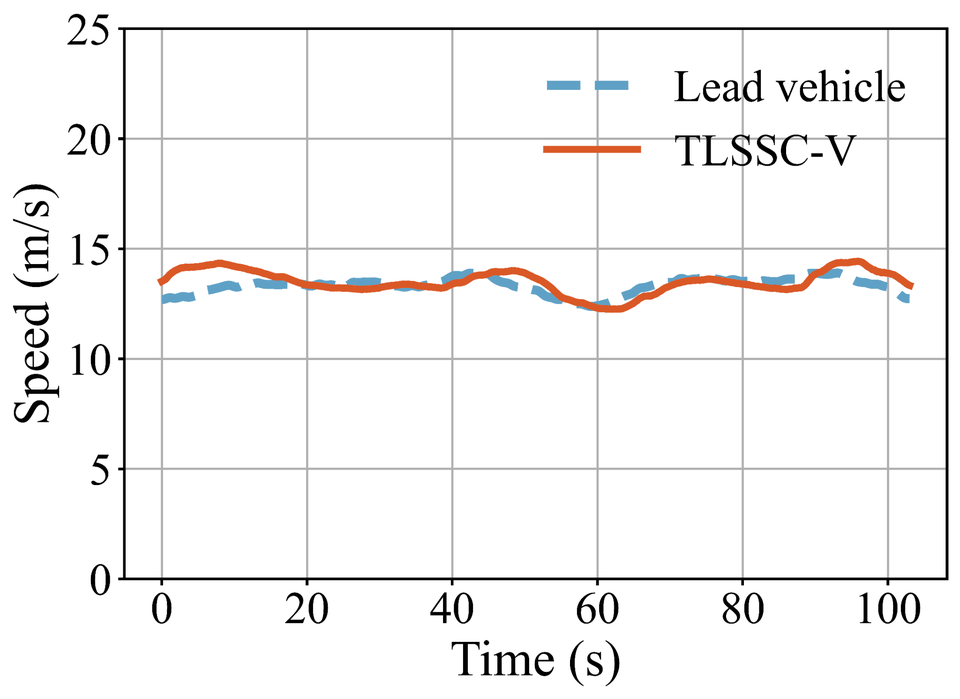}
        \caption{Car-following when proceeding straight through the intersection (Gap level 2)}
    \end{subfigure}
    \hfill
    \begin{subfigure}[t]{0.3\textwidth}
        \centering
        \includegraphics[height=4cm]{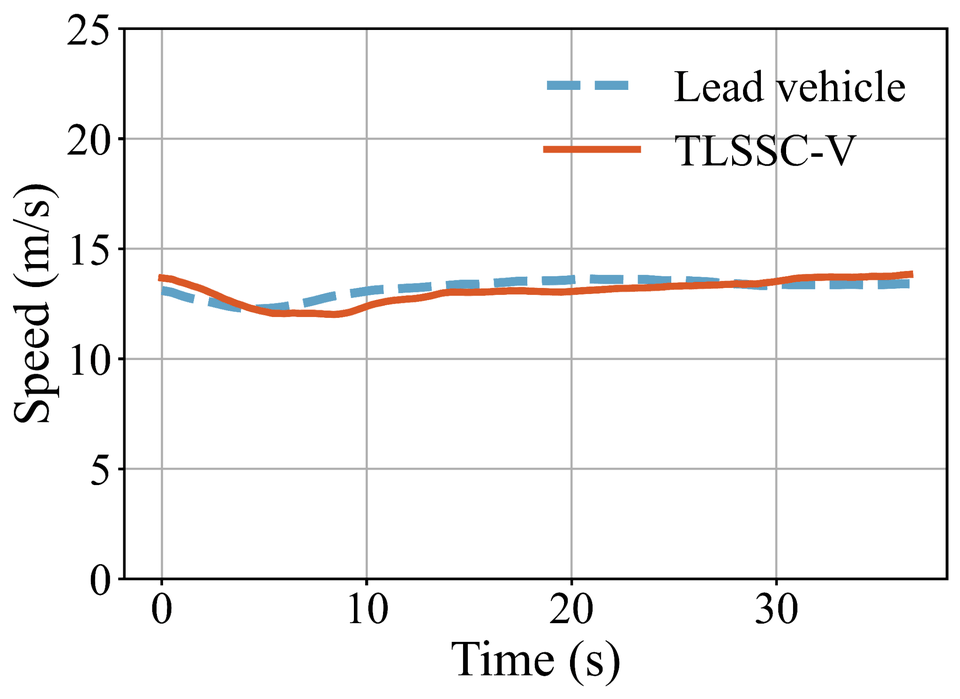}
        \caption{Car-following when proceeding straight through the intersection (Gap level 4)}
    \end{subfigure}
    \hfill
    \begin{subfigure}[t]{0.3\textwidth}
        \centering
        \includegraphics[height=4cm]{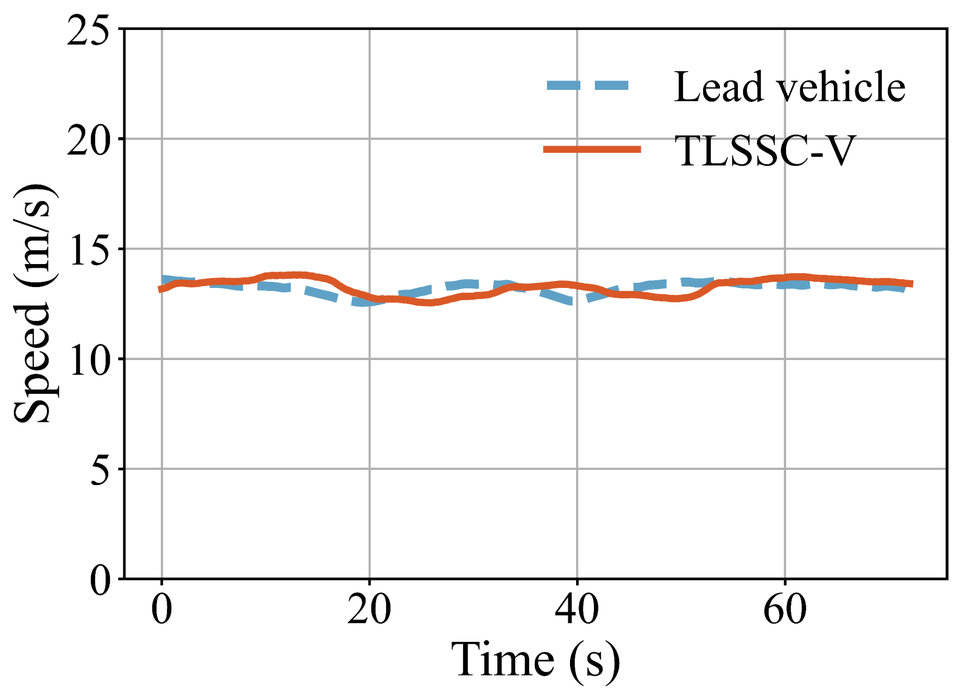}
        \caption{Car-following when proceeding straight through the intersection (Gap level 7)}
    \end{subfigure}    

    \begin{subfigure}[t]{0.3\textwidth}
        \centering
        \includegraphics[height=4cm]{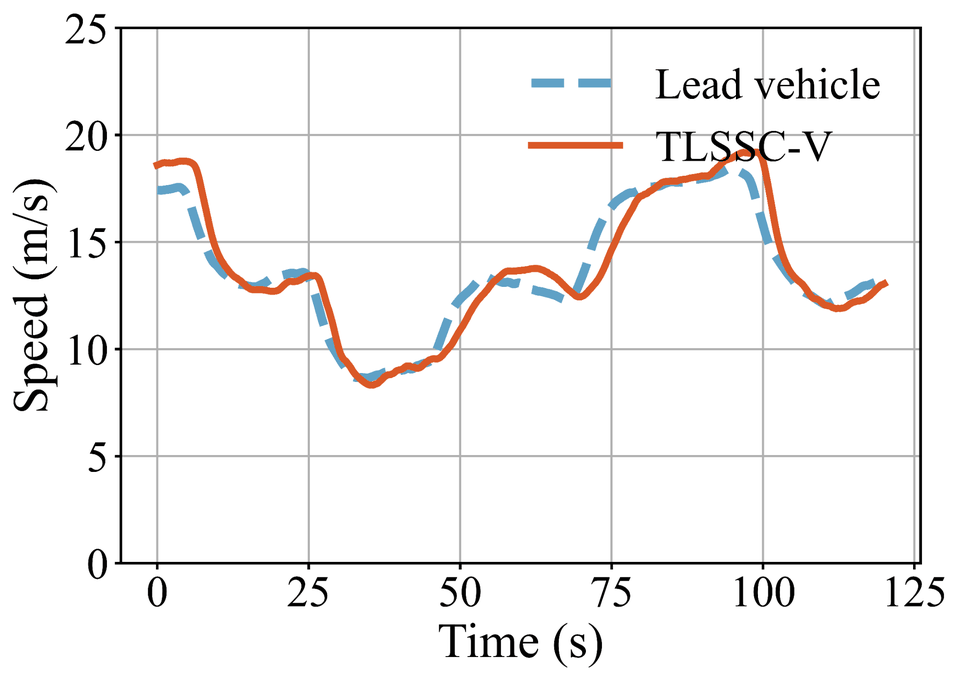}
        \caption{Standard car-following (Gap level 2)}
    \end{subfigure}
    \hfill
    \begin{subfigure}[t]{0.3\textwidth}
        \centering
        \includegraphics[height=4cm]{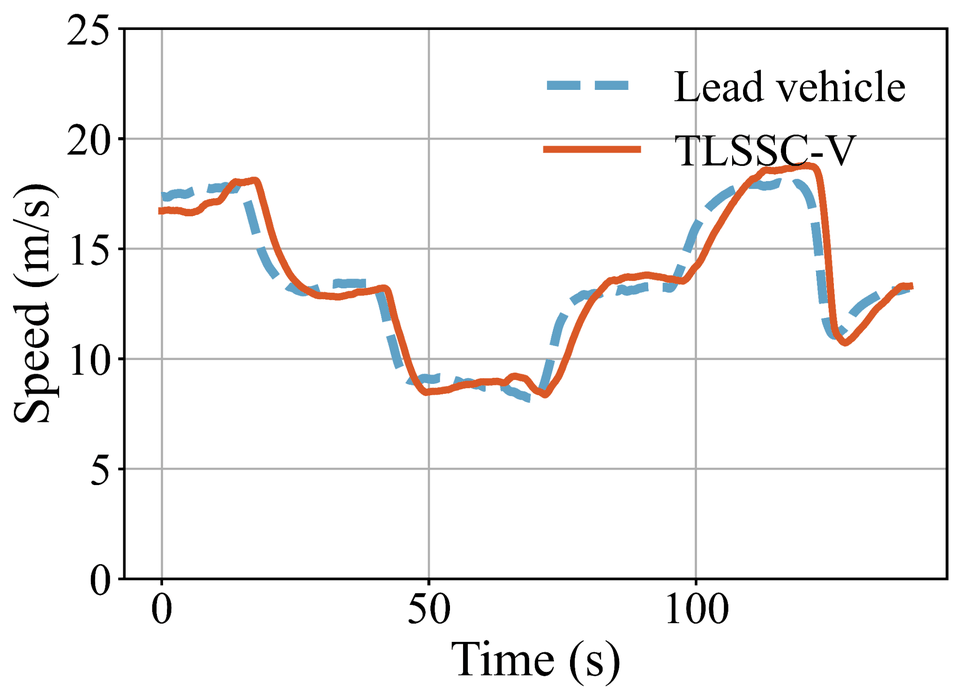}
        \caption{Standard car-following (Gap level 4)}
    \end{subfigure}
    \hfill
    \begin{subfigure}[t]{0.3\textwidth}
        \centering
        \includegraphics[height=4cm]{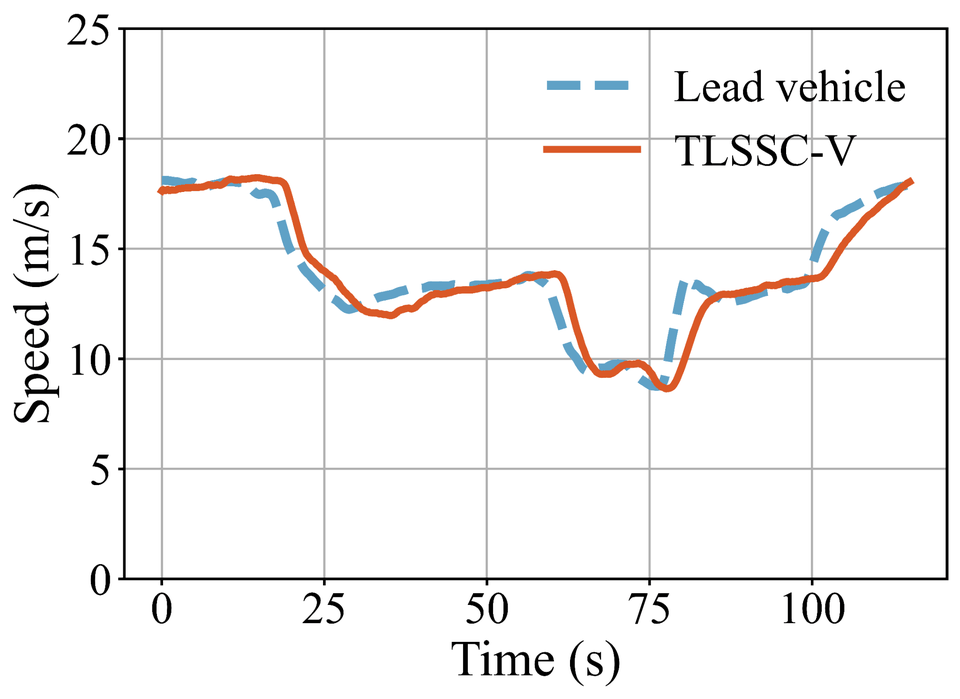}
        \caption{Standard car-following (Gap level 7)}
    \end{subfigure}
    
    \caption{Examples of speed profiles in the TLSSC dataset.}
    \label{fig:speed profile}
\end{figure}

 Figure \ref{fig:speed profile} presents examples of speed profiles from the TLSSC dataset. Figure \ref{fig:speed profile} (a)–(c) illustrate stopping behaviors, while Figure \ref{fig:speed profile} (d)–(f) show accelerating behaviors. Figure \ref{fig:speed profile} (g)–(i) depict car-following behaviors when TLSSC-V passes through an intersection under a green light while following a lead vehicle, with different gap level settings. Figure \ref{fig:speed profile} (j)–(l) show examples of standard car-following behaviors under various gap level settings. For stopping and accelerating behaviors, we present the speed profiles of the TLSSC-V. For car-following behaviors, we provide the speed profiles of both the lead vehicle and the TLSSC-V.

\subsection{Data structure}

\par Table \ref{tab:data structure} summarizes the structure of the TLSSC dataset, including the notation used for key variables, their corresponding column names in the data files, and a brief explanation of each. The dataset contains timestamped vehicle trajectory information for both the TLSSC-V and the lead vehicle, including raw and smoothed values for position and speed.

\begin{table}[h!]
\centering
\caption{Data structure for the TLSSC dataset}
\label{tab:data structure}
\begin{tabular}{|>{\centering\arraybackslash}m{3.5cm} 
                |>{\centering\arraybackslash}m{5cm} 
                |>{\raggedright\arraybackslash}m{6.5cm}|}
\hline
\textbf{Notation} & \textbf{Column name} & \textbf{Explanation} \\
\hline
\(t, \forall t \in \mathbb{N^{+}} \) & Time & ISO 8601 format time record, which includes the full date, time, and time zone offset. \\
\hline
\multirow{2}{*}{\(x_t, y_t, \forall t \in \mathbb{N^+}\)} 
& Longitude, Latitude (or Longitude\_follow, Latitude\_follow)  
& The longitude and latitude coordinates of the TLSSC-V at the timestep \(t\). \\
\cline{2-3}
& Longitude\_smoothed, Latitude\_smoothed (or Longitude\_follow\_smoothed, Latitude\_follow\_smoothed)
& The longitude and latitude coordinates of the TLSSC-V at the timestep \(t\) after moving average smoothing. \\
\hline
\multirow{2}{*}{\(v_t, \forall t \in \mathbb{N^+}\)} & Speed & The speed of the TLSSC-V at the timestep \(t\).  The unit is \(m/s\).\\
\cline{2-3}
& Speed\_smoothed & The speed of the TLSSC-V at the timestep \(t\) after moving average smoothing. The unit is \(m/s\).\\
\hline
\multirow{2}{*}{\(x_t^{\text{lead}}, y_t^{\text{lead}}, \forall t \in \mathbb{N^+}\)} 
& Longitude\_lead, Latitude\_lead  
& The longitude and latitude coordinates of the lead vehicle at the timestep \(t\). \\
\cline{2-3}
& Longitude\_lead\_smoothed, Latitude\_lead\_smoothed 
& The longitude and latitude coordinates of the lead vehicle at the timestep \(t\) after moving average smoothing. \\
\hline
\multirow{2}{*}{\(v_t^{\text{lead}}, \forall t \in \mathbb{N^+}\)} & Speed\_lead & The speed of the lead vehicle at the timestep \(t\). The unit is \(m/s\).\\
\cline{2-3}
& Speed\_lead\_smoothed & The speed of the lead vehicle at the timestep \(t\) after moving average smoothing. The unit is \(m/s\).\\
\hline
\end{tabular}
\end{table}

\par In addition to the commonly used kinematic variables listed in Table \ref{tab:data structure}, the TLSSC dataset also includes several supplementary attributes that provide additional context for spatial accuracy and measurement reliability. These include elevation (altitude relative to the Earth's ellipsoid surface), instrument height (height of the GPS sensor above the mounting point), and bearing (the vehicle's heading direction in degrees from true north). To assess positioning precision, the TLSSC dataset contains horizontal and vertical accuracy estimates (in meters), as well as dilution of precision metrics: PDOP (position), HDOP (horizontal), and VDOP (vertical). These values help users evaluate the quality of spatial data and make informed decisions in downstream trajectory analysis tasks.

\subsection{Quality assessment results}
Table \ref{tab: quality assessment summary} summarizes the trajectory data collected in the TLSSC dataset, categorized into three behavior types: stopping, accelerating, and car-following. For each TLSSC behavior, it reports the number of trajectory segments, total travel distance, duration, and the percentage of anomalous acceleration and jerk values before and after smoothing. Car-following behaviors dominate in both distance (25,406.48 m) and duration (1,981.10 s) in the TLSSC dataset, reflecting their longer continuous sequences, while stopping and accelerating behaviors contribute 6,418.25 m and 3,104.58 m, respectively. The anomaly metrics demonstrate that the smoothing process effectively eliminated noise, reducing acceleration and jerk anomalies to near zero. Overall, the dataset contains 74 segments covering over 41 km and 3,409 seconds, with quality metrics confirming that the data are clean and reliable for further TLSSC behavioral analysis.

\begin{table}[h!]
\centering
\caption{Trajectory segment quantity, distance, duration, and quality assessment results for each TLSSC behavior in TLSSC dataset.}
\label{tab: quality assessment summary}
\begin{tabular}{|>{\centering\arraybackslash}p{3.8cm} 
                |>{\centering\arraybackslash}p{2cm} 
                |>{\centering\arraybackslash}p{2cm} 
                |>{\centering\arraybackslash}p{2cm} 
                |>{\centering\arraybackslash}p{2cm} 
                |>{\centering\arraybackslash}p{2cm}|}
\hline
Behavior & Trajectory segments quantity & Distance (m) & Duration (s) & Anomaly Acceleration (\%) & Anomaly Jerk (\%)\\
\hline
Stopping behaviors & 30 & 6,418.25 & 582.80 & 0.17 / 0.00 & 1.11 / 0.00\\
\hline
Accelerating behaviors & 28 & 3,104.58 & 317.90 & 0.24 / 0.00 & 1.39 / 0.00\\
\hline
Car-following behaviors & 31 & 25,406.48 & 1,981.10 & 0.00 / 0.00 & 0.47 / 0.00 \\
\hline
All behaviors & 74   & 41,037.56   & 3,409.50  & 0.07 / 0.01 & 0.84 / 0.03 \\
\hline

\end{tabular}

\vspace{0.5em}
\begin{flushleft}
\footnotesize
\textit{Note:} The numbers before and after the slash represent the trajectory data quality assessment results before and after enhancement, respectively. For example, “0.17 / 0.00” means that the trajectory data quality assessment result is 0.17 before enhancement and 0.00 after enhancement.
\end{flushleft}
\end{table}

\section{Behavior analysis}
\par This section presents an analysis of TLSSC behaviors based on the collected TLSSC dataset. We first introduce the model formulation established to characterize TLSSC behaviors, followed by the modeling results and an analysis of the observed behaviors.

\subsection{Behavior model formulation}
\par We modeled the behavior of the TLSSC using a car-following model known as the FVDM (\cite{jiang2001full}). FVDM is an extended car-following model developed from the classical Optimal Velocity Model (OVM), with improved stability and realism. It introduces a speed difference term to capture the influence of speed variations between lead and following vehicles, thereby enhancing the model's ability to reflect complex car-following behavior (\cite{yu2013full}).

\par The acceleration of the following vehicle (TLSSC-V in this work) at time $t$ is calculated through Equations \ref{eq:fvdm main}-\ref{eq:fvdm optimzal speed} in FVDM.

\begin{equation}
    \frac{dv_n(t)}{dt} = \alpha \left[ V(s_n(t)) - v_n(t) \right] + \beta \left[ v_{n-1}(t) - v_n(t) \right]
    \label{eq:fvdm main}
\end{equation}
\medskip
\begin{equation}
    s_t = x_t^{lead} - x_t
    \label{eq:fvdm spacing}
\end{equation}
\medskip

\noindent where:

\par $v_t$: speed of the following vehicle at time $t$,
\par $v^{lead}_t$: speed of the lead vehicle at time $t$,
\par $x_t$: longitudinal position of the following vehicle at time $t$,
\par $x^{lead}_t$: longitudinal position of the lead vehicle at time $t$,    
\par $s_t$: spacing between lead and following vehicle,
\par $\alpha$: sensitivity coefficient to the optimal speed,
\par $\beta$: sensitivity coefficient to the speed difference,
\par $V(s)$: optimal speed function, defining the desired speed based on spacing.

\par The optimal speed function $V(s)$ is typically modeled as a smooth increasing function with a saturation effect as shown in Equation \ref{eq:fvdm optimzal speed}.

\begin{equation}
    V(s) = v_{\max} \cdot \tanh\left( \frac{s - s_0}{\Delta s} \right)
    \label{eq:fvdm optimzal speed}
\end{equation}

\noindent where:
\par $v_{\max}$: maximum desired speed,
\par $s_0$: minimum desired spacing,
\par $\Delta s$: sensitivity spacing parameter that controls the steepness of the transition.

\par The form of the FVDM ensures that the desired speed approaches zero when the spacing is small, and asymptotically approaches $v_{\max}$ as the spacing becomes large. The hyperbolic tangent shape of $V(s)$ enables a gradual and realistic transition between congested and free-flow regimes.

\par Equations \ref{eq:fvdm main}–\ref{eq:fvdm optimzal speed} are used to model both the standard car-following behavior and the car-following behavior when proceeding straight through an intersection, where the trajectories of the lead vehicle and the TLSSC-V can be incorporated to fit the car-following parameters. For stopping and accelerating behaviors, we extend Equations \ref{eq:fvdm main}–\ref{eq:fvdm optimzal speed} with the following modifications to better capture the dynamics of the stopping and accelerating.

\par For accelerating behavior, we assume that \( v^{\text{lead}}_t = v_{\max} \) and \( x^{\text{lead}}_t = +\infty \), meaning the lead vehicle of the TLSSC-V is positioned infinitely far ahead. In this case, the gap between the two vehicles is extremely large, and the TLSSC-V is not influenced by the lead vehicle, thus accelerating toward its maximum speed. For stopping behavior, we assume that \( v^{\text{lead}}_t = 0 \) and \( x^{\text{lead}}_t = x^{\text{stopline}} \), indicating that the lead vehicle is stationary at the stop line. The TLSSC-V will follow this lead vehicle and come to a stop in front of the stop line.

\subsection{Model parameters calibration}

Following four parameters of the FVDM are calibrated using the TLSSC dataset: $\alpha$, controls the strength of the vehicle's adjustment to its optimal speed; $\beta$, regulates the response to the speed difference with the lead vehicle; $s_0$, defines the threshold spacing below which desired speed is significantly reduced; $\Delta s$, influences how quickly the optimal speed increases with spacing. For \( v_{\max} \), which determines the upper bound of the optimal speed function, we did not calibrate this parameter in this work. Instead, we set its value to the maximum speed assigned to the TLSSC-V in the TLSSC dataset to ensure consistency between the model and the observed behavior in the data.

We used the global optimization algorithm DIviding RECTangles (DIRECT) to optimize the parameters based on the data from the TLSSC dataset. DIRECT is a derivative-free global optimization algorithm that iteratively partitions the parameter space and balances global exploration with local refinement (\cite{hansen1995lipschitz}). Its ability to efficiently handle non-linear and multi-modal relationships makes it well-suited for calibrating driving behavior models. During the search for optimal parameter values, we chose to minimize the RMSE between the model-predicted speed and the observed speed in the TLSSC dataset as the objective function. The parameter bounds were set as follows: \( \alpha \in [0, 5] \), \( \beta \in [0, 5] \), \( s_0 \in [0, 10] \), and \( \Delta \in [0.1, 20] \).

\subsection{Behavior model calibration results}

\par The calibrated FVDM model parameters for all behaviors, along with their corresponding speed RMSE values, are summarized in Table \ref{tab: model calibration results}. Based on the calibrated FVDM parameters presented in Table~\ref{tab: model calibration results}, we observe distinct behavioral characteristics across the different TLSSC behaviors

\begin{table}
    \caption{FVDM calibration results.}
    \centering
    \begin{tabular}{|>{\centering\arraybackslash}m{5.5cm} 
                    |>{\centering\arraybackslash}m{1.7cm} 
                    |>{\centering\arraybackslash}m{1.7cm} 
                    |>{\centering\arraybackslash}m{1.7cm} 
                    |>{\centering\arraybackslash}m{1.7cm} 
                    |>{\centering\arraybackslash}m{1.7cm}|}
        \hline
        Behavior & $\alpha$ & $\beta$ & $s_0$ & $\Delta s$ & Speed RMSE (m/s) \\ \hline
        Stopping behavior & 0.7510 & 0.8127 & 5.5761 & 18.9590 & 1.6716\\ \hline
        Accelerating behavior & 0.0926 & 0.0926 & 5.0000 & 18.8944 & 1.2359\\ \hline         
        Standard car-following behavior (Gap level 2) & 0.0309 & 1.6770 & 8.8272 & 13.4895 & 0.9252\\ \hline
        Standard car-following behavior (Gap level 4) & 0.0171 & 3.3368 & 9.9108 & 19.7680 & 0.9252\\ \hline
        Standard car-following behavior (Gap level 7) & 0.0309 & 3.4259 & 9.8148 & 19.6315 & 0.9252\\ \hline
        Car-following behavior when proceeding straight through the intersection (Gap level 2) & 0.0034 & 1.6701 & 6.6735 & 3.0618 & 0.3591 \\ \hline
        Car-following behavior when proceeding straight through the intersection (Gap level 4) & 0.0926 & 0.2160 & 4.2593 & 10.5414 & 0.3322 \\ \hline
        Car-following behavior when proceeding straight through the intersection (Gap level 7) & 0.0103 & 0.1680 & 9.2524 & 10.4049 & 0.2514 \\ \hline
    \end{tabular}
    \label{tab: model calibration results}
\end{table}

\subsubsection{Stopping behavior}
\par The stopping behavior shows relatively high values of both $\alpha = 0.7510$ and $\beta = 0.8127$, indicating that the vehicle strongly and sensitively adjusts its speed. The threshold spacing $s_0 = 5.5761$\,m and the spacing growth factor $\Delta s = 18.9590$\,m suggest that the optimal speed increases steadily as spacing increases. These results indicate a proactive and cautious deceleration strategy, where TLSSC-V begins slowing down early in response to red lights or stop signs.

\subsubsection{Accelerating behavior}
\par In contrast, accelerating behavior shows low values of $\alpha = 0.0926$ and $\beta = 0.0926$, indicating that the vehicle has a mild tendency to adjust its speed toward the optimal value. The $s_0 = 5.0000$\,m and $\Delta s = 18.8944$\,m remain comparable to the stopping behavior, suggesting similar spacing dynamics. However, the vehicle accelerates more conservatively and gradually, which is consistent with the permission-based nature of the TLSSC where the human driver input is required before resuming movement.

\subsubsection{Car-following behavior}
\par For standard car-following behaviors under increasing gap levels (2, 4, and 7), the calibrated $\beta$ values grow significantly (from 1.6770 to 3.4259), while $\alpha$ remains low (0.0171 to 0.0309), implying that TLSSC-V mainly reacts to speed differences with the lead vehicle rather than its own deviation from optimal speed. The $s_0$ and $\Delta s$ values also increase slightly with gap level, suggesting that the vehicle maintains larger headways and adapts its optimal speed more cautiously. The consistent speed RMSE of 0.9252\,m/s across gap levels indicates reliable model performance in standard car-following scenarios.

\par However proceeding straight through intersections displays distinct behavior. At gap level 2, $\alpha$ is extremely low (0.0034), and $\beta = 1.6701$ is relatively high, meaning the vehicle relies almost entirely on closing speed for control. As the gap level increases, both $\alpha$ and $\beta$ decrease sharply, showing reduced responsiveness and greater inertia in control. The threshold spacing $s_0$ and spacing growth factor $\Delta s$ are generally lower than those in standard following, reflecting tighter headways and quicker transitions in optimal speed. The speed RMSEs in this category are the lowest among all behaviors (as low as 0.2514\,m/s), suggesting more stable dynamics when traversing intersections under green signals.

\par In summary, the FVDM parameters capture nuanced differences across TLSSC-V behavior types. Stopping involves sharp and early braking, acceleration is conservative and gradual, standard car-following is primarily governed by relative speed under large gaps, and intersection-following demonstrates even smoother, less aggressive dynamics.

\subsection{Car-following threshold insights}

\par While observing TLSSC behavior, we identified an insightful phenomenon regarding how the TLSSC-V responds to green lights based on the presence of a lead vehicle. Specifically, when a lead vehicle is detected ahead, TLSSC-V follows it through the intersection without decelerating or requiring permission from the human driver. In contrast, when no lead vehicle is detected, TLSSC-V slows down as it approaches the green light and requests permission from the human driver. If permission is not granted, the vehicle comes to a complete stop at the stop line. This raises an interesting question: at what distance does a lead vehicle need to be for TLSSC-V to detect it and initiate a car-following response instead of a stopping response? We define this distance as the car-following threshold, which is crucial because it determines whether TLSSC-V adopts stopping behavior or car-following behavior in response to a detected green light.

\par To estimate this threshold, we conducted additional experiments based on the earlier car-following setup, using a speed setting of 40 mph and gap level 7. In each trial, the lead vehicle cruised at a constant speed along the approach lane toward an intersection, and the human driver activated the TLSSC feature at specific distances behind the lead vehicle approximately 40 m, 60 m, 90 m, and 150 m. The goal was to observe whether TLSSC-V would detect the lead vehicle and follow it through the green light or slow down and stop. It is important to note that this threshold estimation is approximate due to limitations in precisely controlling inter-vehicle distance at the moment of TLSSC activation by human driver.

\begin{figure}[htbp]
    \centering
    \begin{subfigure}{0.32\textwidth}
    \centering
        \includegraphics[width=\linewidth]{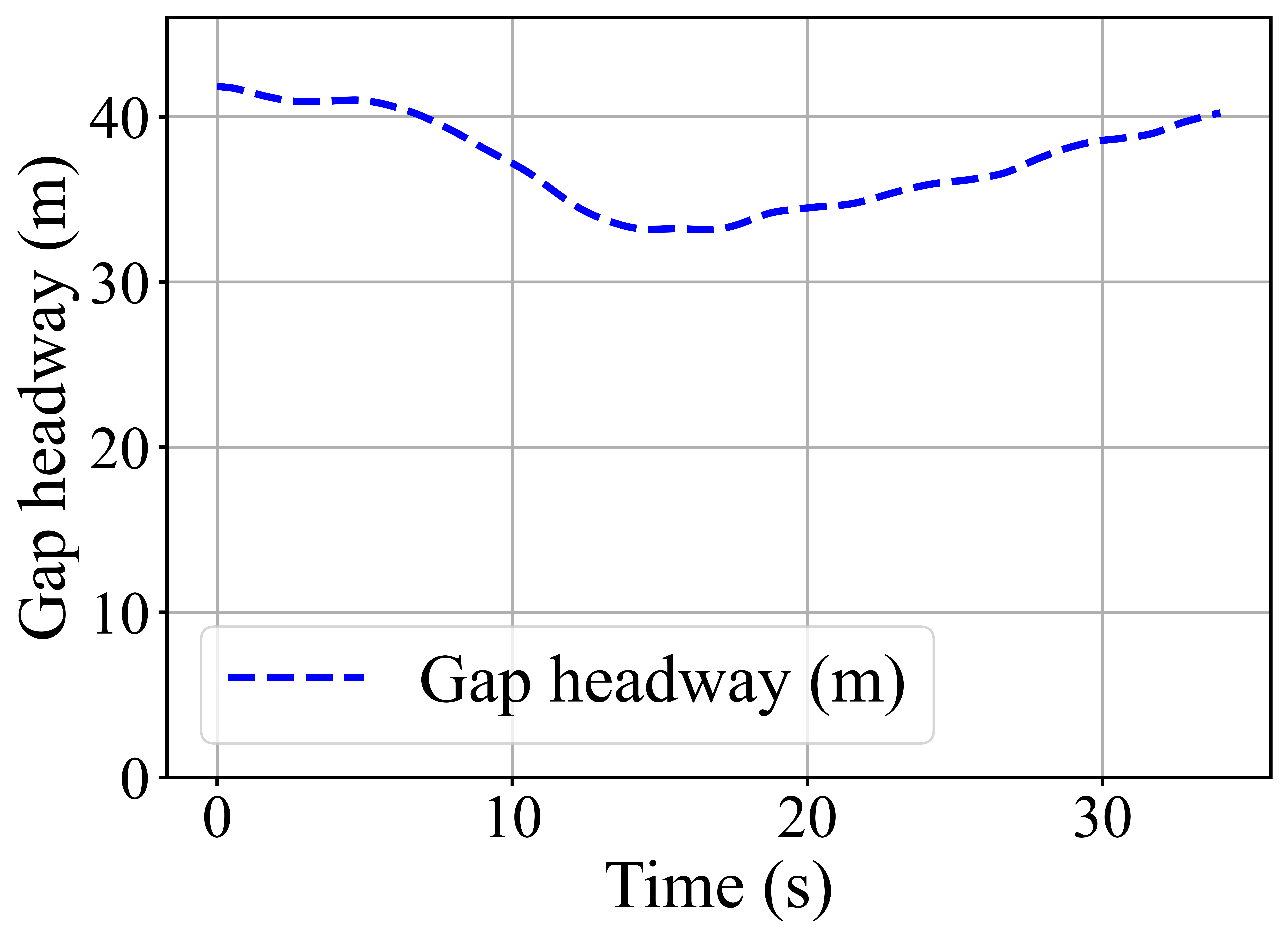}
        \caption{Gap headway 1.}
    \end{subfigure}
    \hfill
    \begin{subfigure}{0.32\textwidth}
    \centering
        \includegraphics[width=\linewidth]{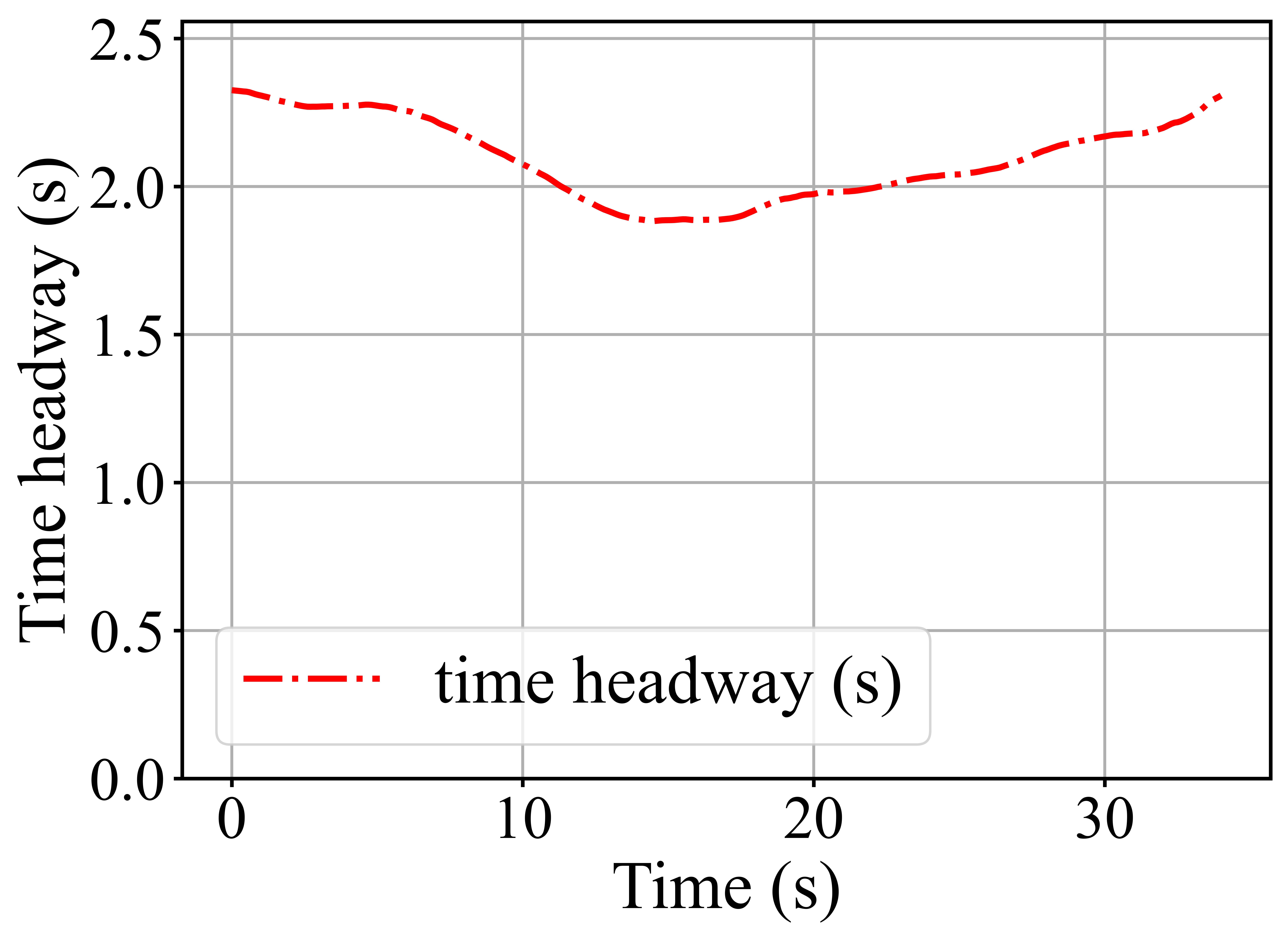}
        \caption{Time headway 1.}
    \end{subfigure}
    \hfill
    \begin{subfigure}{0.32\textwidth}
    \centering
        \includegraphics[width=\linewidth]{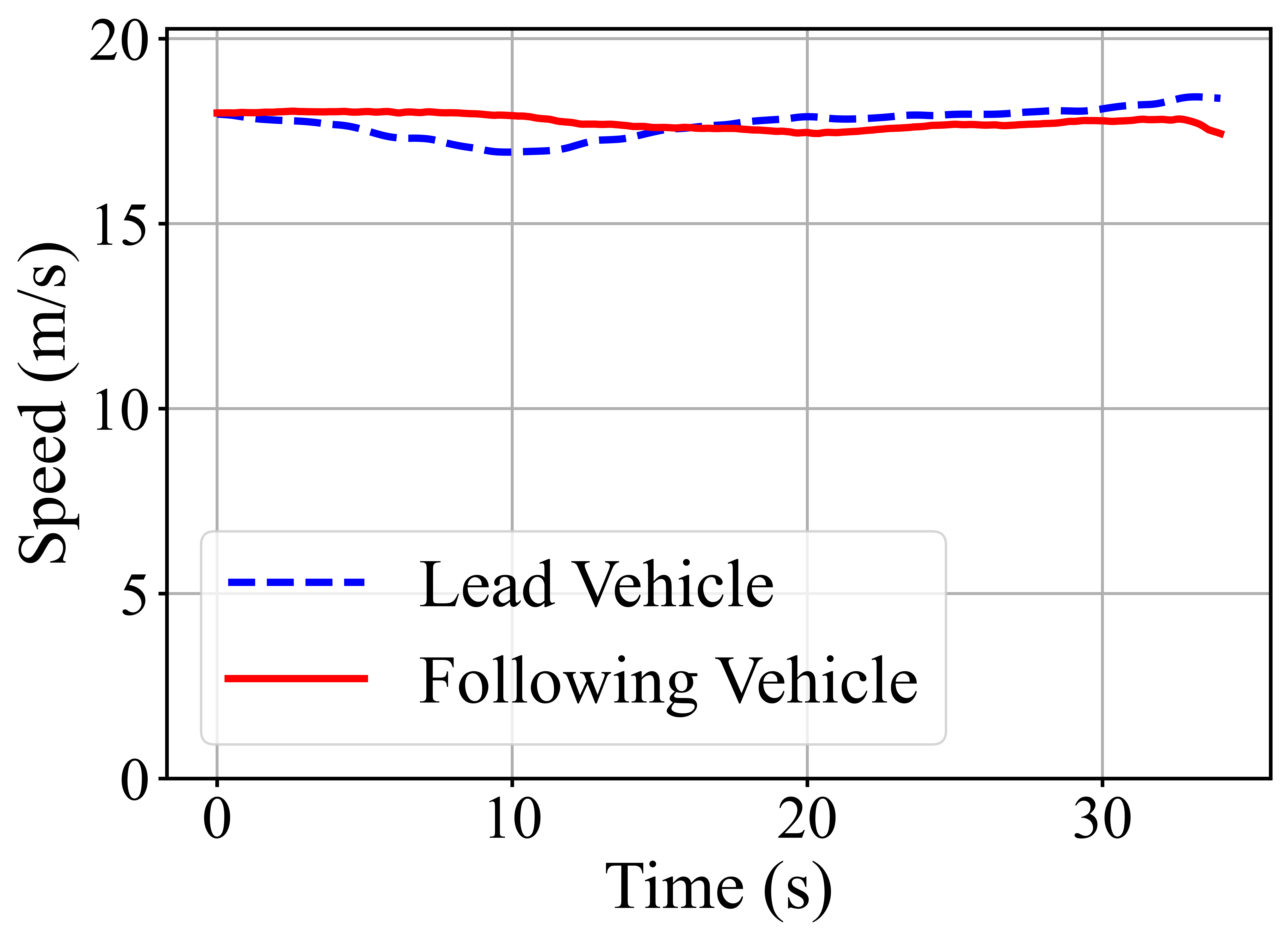}
        \caption{Speed 1.}
    \end{subfigure}
    
    \begin{subfigure}{0.31\textwidth}
    \centering
        \includegraphics[width=\linewidth]{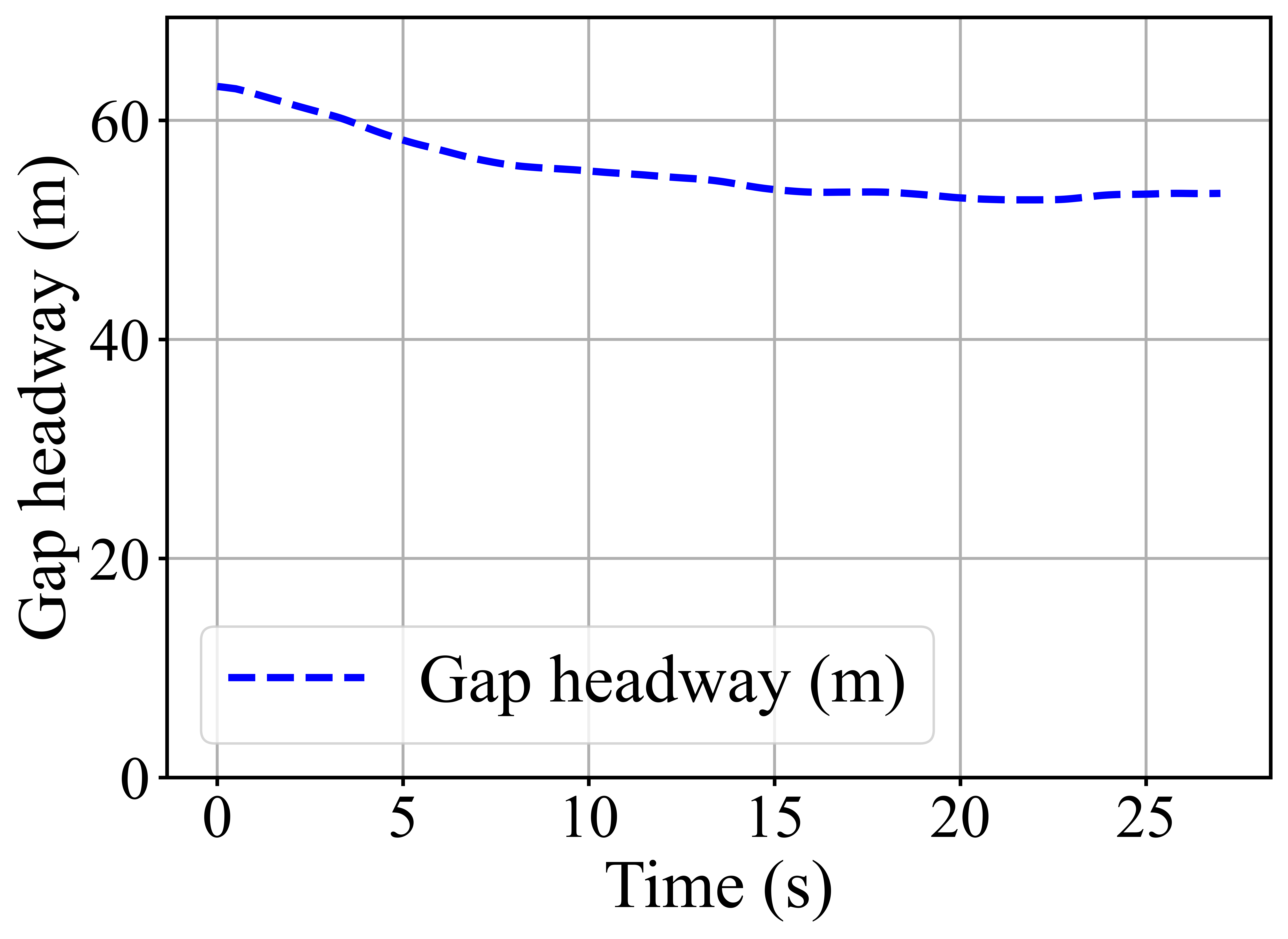}
        \caption{Gap headway 2.}
    \end{subfigure}
    \hfill
    \begin{subfigure}{0.31\textwidth}
    \centering
        \includegraphics[width=\linewidth]{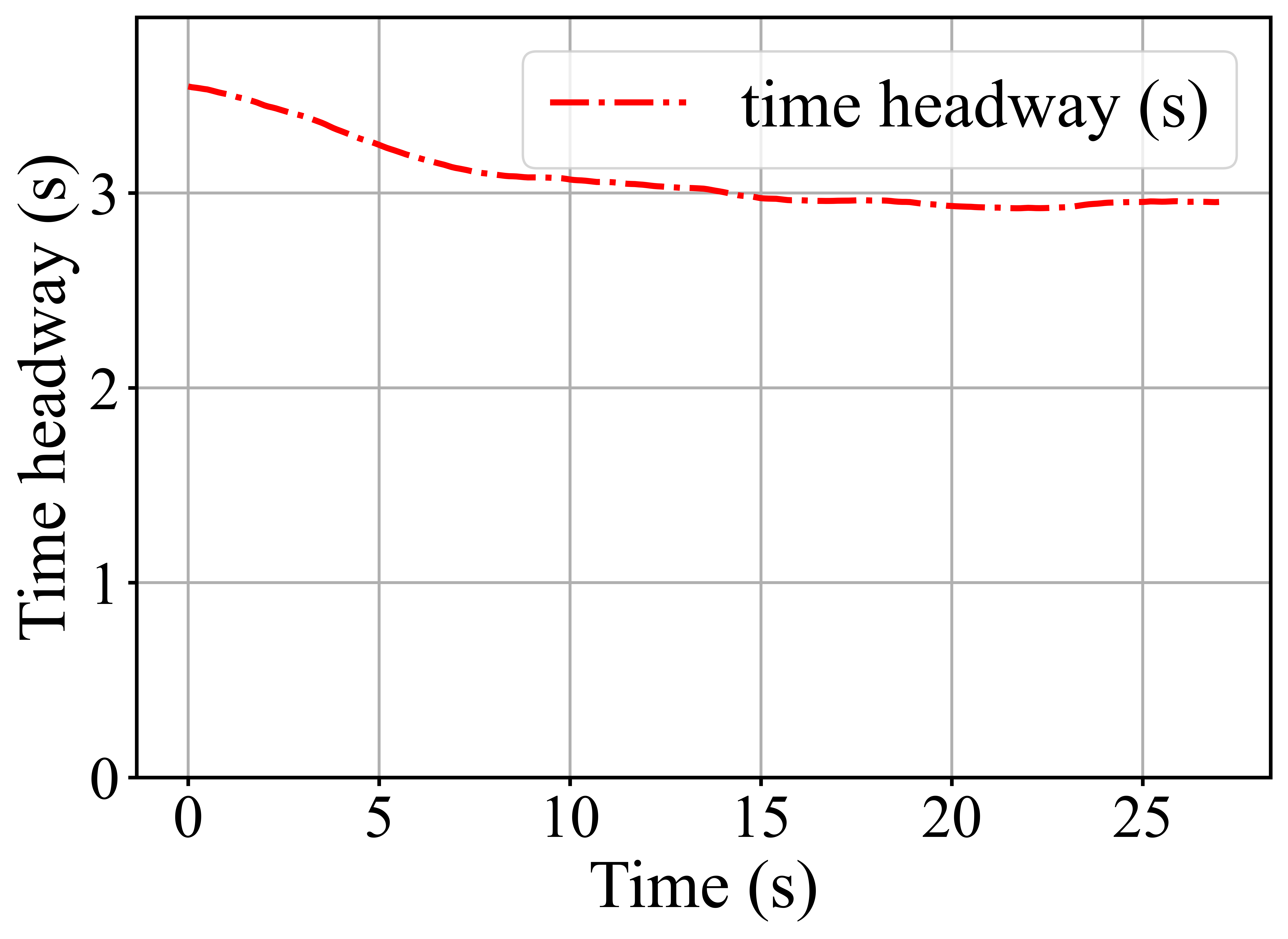}
        \caption{Time headway 2.}
    \end{subfigure}
    \hfill
    \begin{subfigure}{0.31\textwidth}
    \centering
        \includegraphics[width=\linewidth]{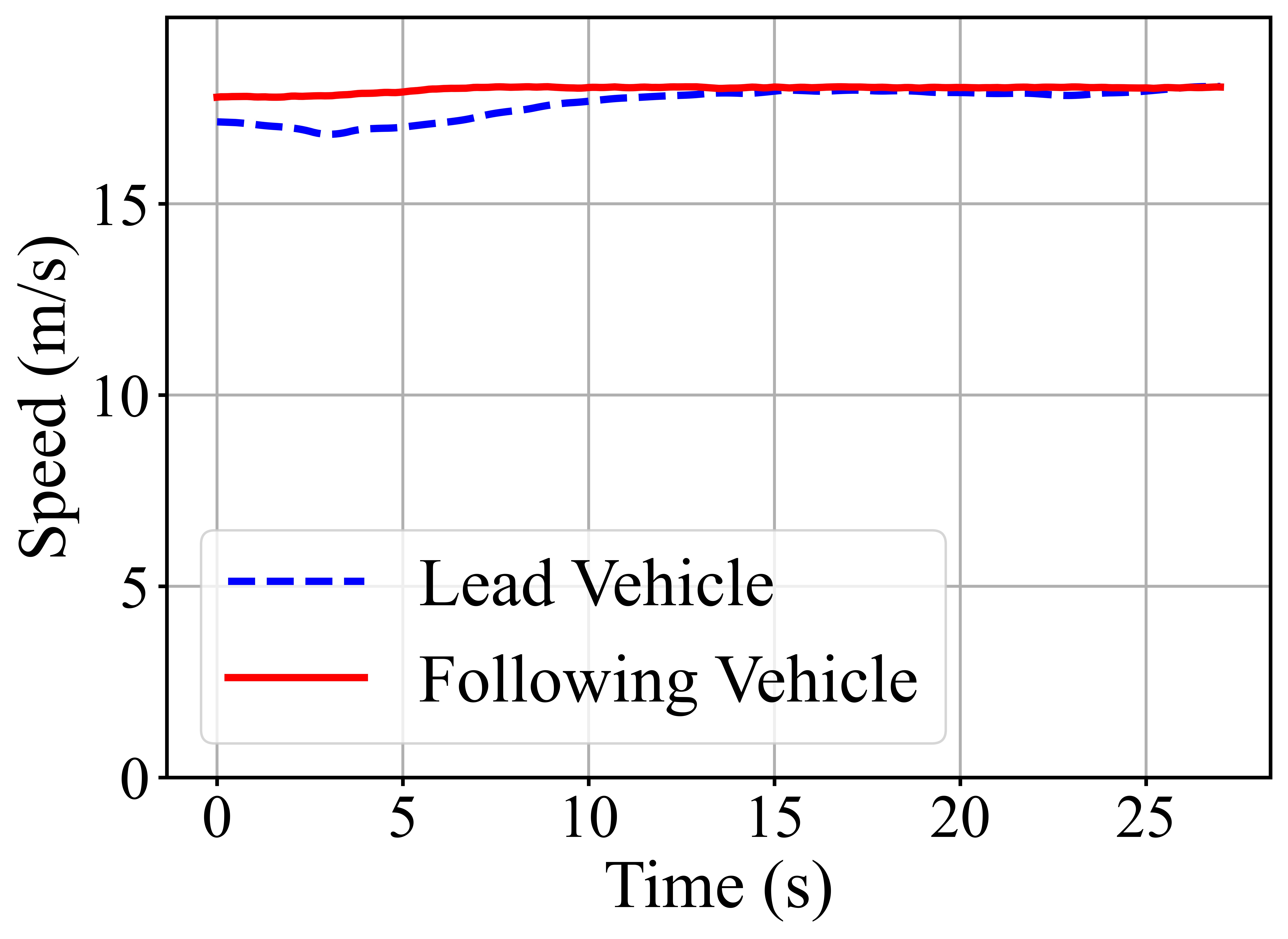}
        \caption{Speed 2.}
    \end{subfigure}
    
    \begin{subfigure}{0.31\textwidth}
    \centering
        \includegraphics[width=\linewidth]{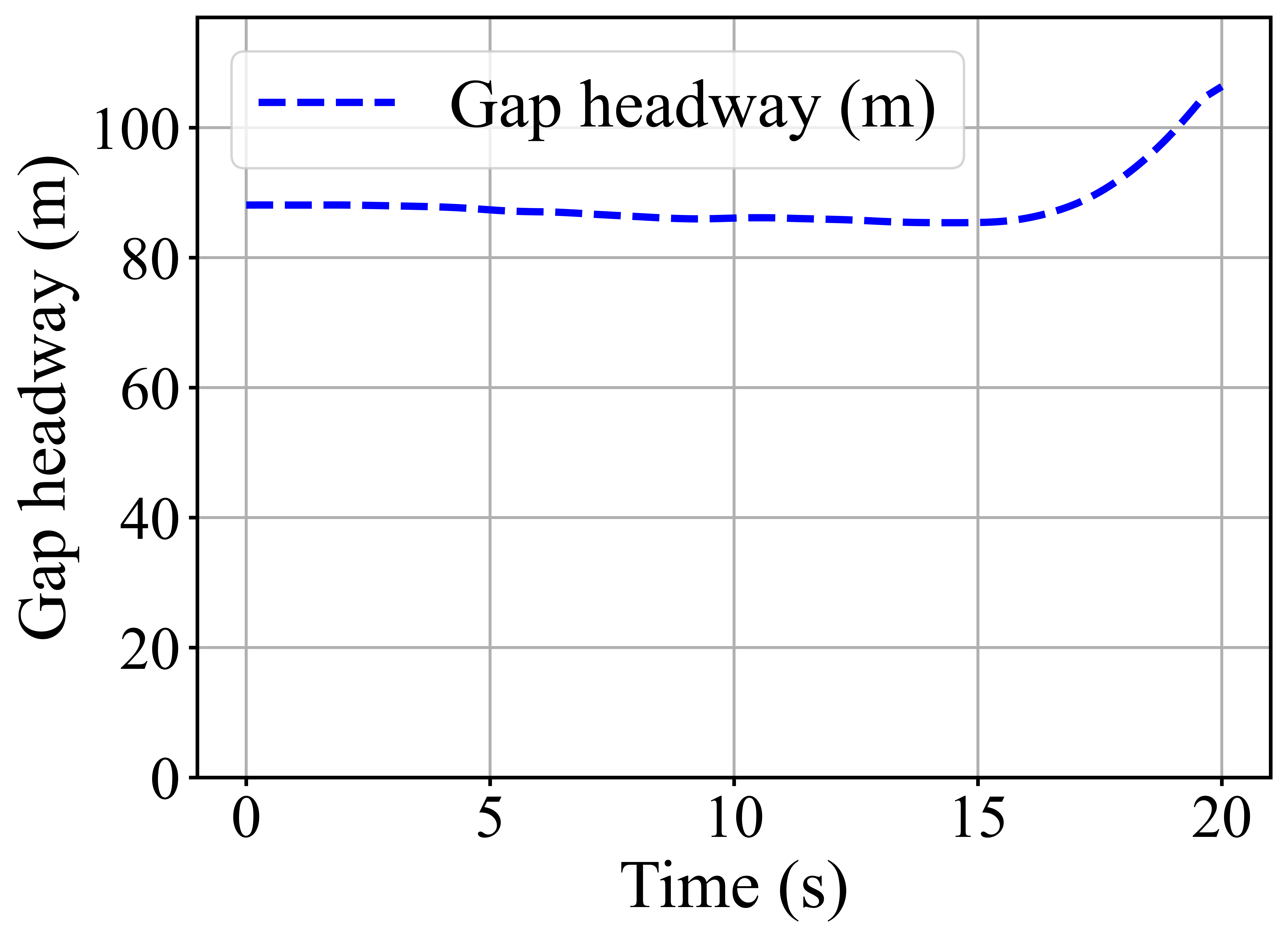}
        \caption{Gap headway 3.}
    \end{subfigure}
    \hfill
    \begin{subfigure}{0.31\textwidth}
    \centering
        \includegraphics[width=\linewidth]{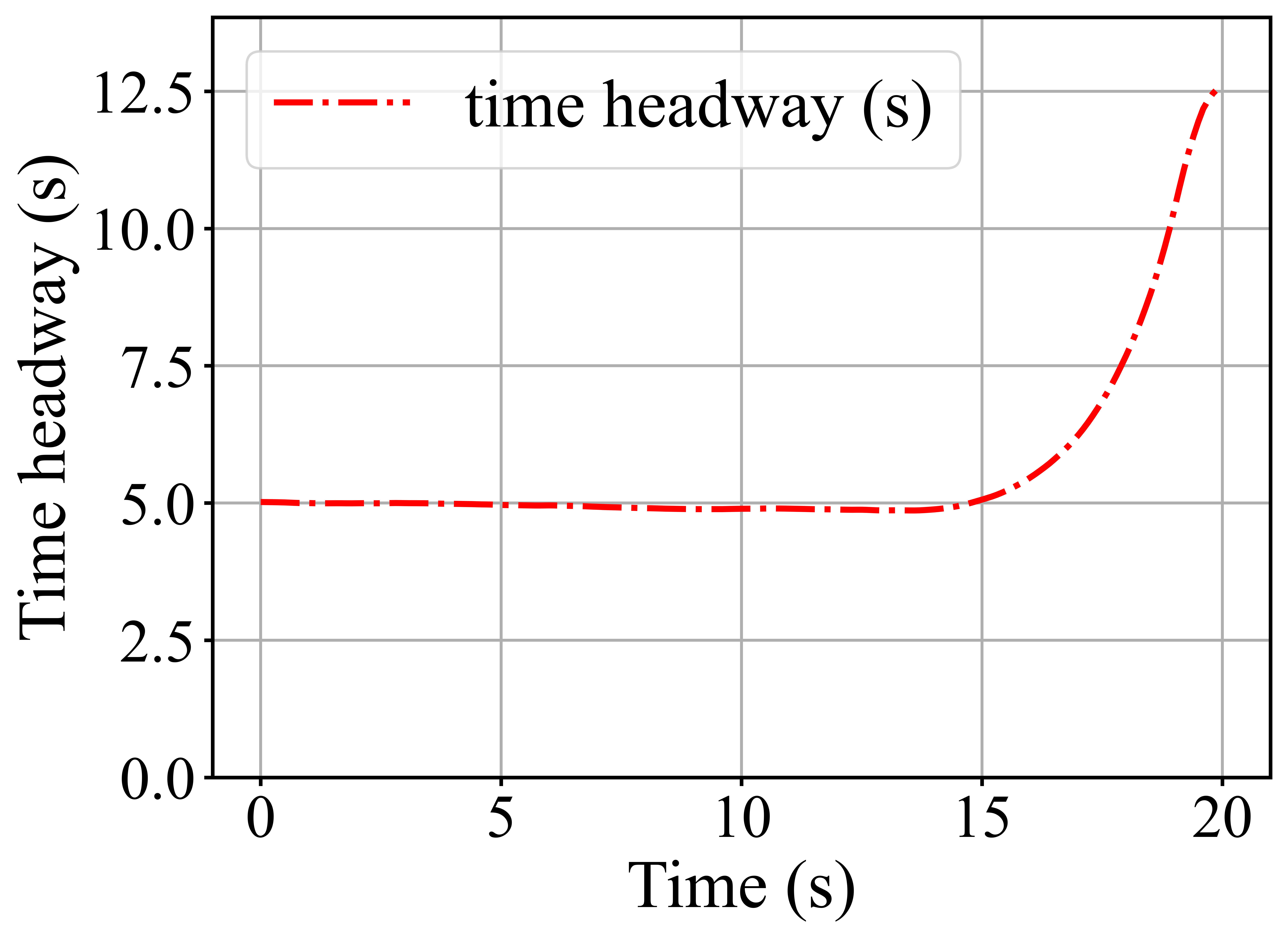}
        \caption{Time headway 3.}
    \end{subfigure}
    \hfill
    \begin{subfigure}{0.31\textwidth}
    \centering
        \includegraphics[width=\linewidth]{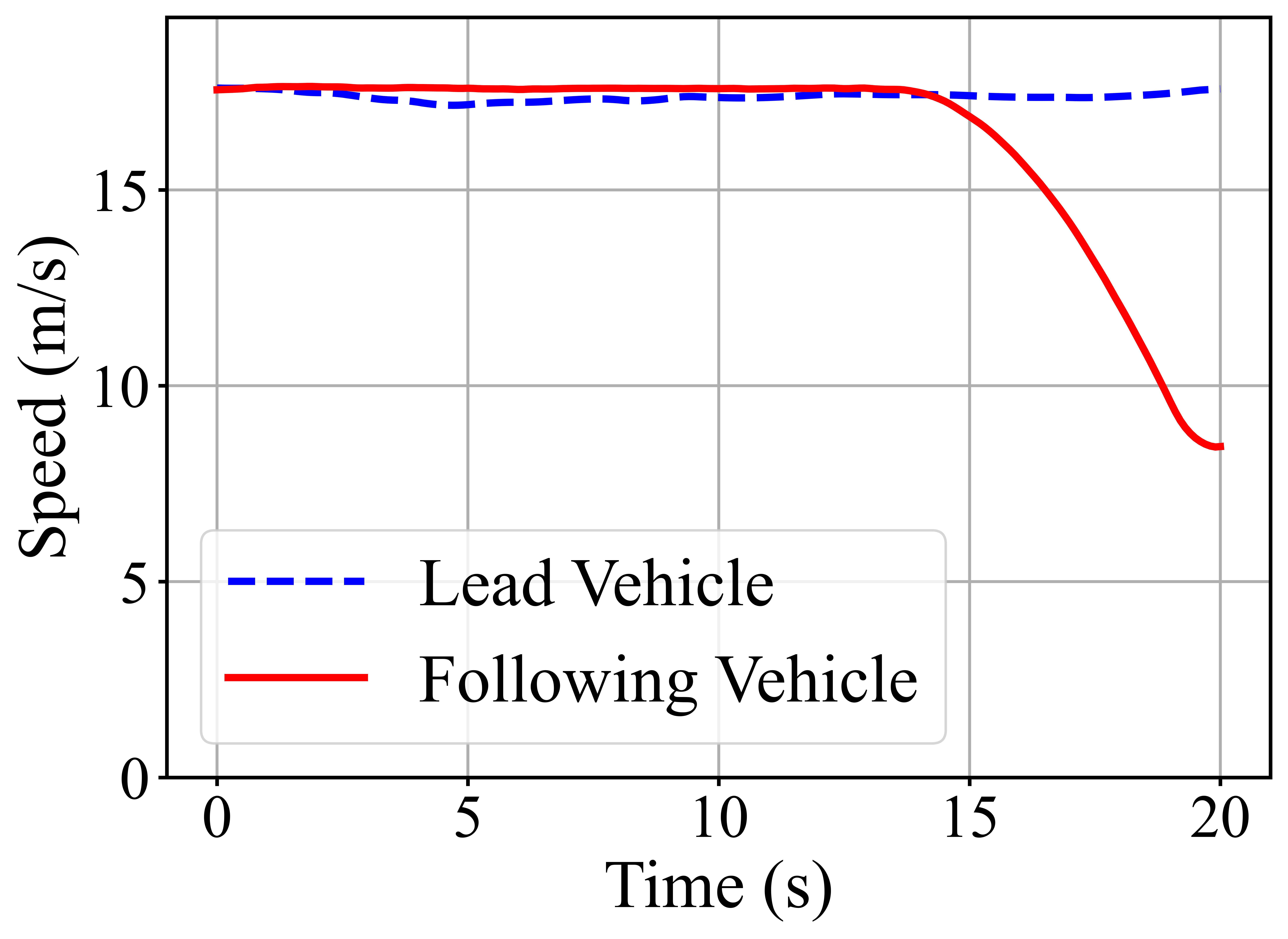}
        \caption{Speed 3.}
    \end{subfigure}
    
    \begin{subfigure}{0.31\textwidth}
    \centering
        \includegraphics[width=\linewidth]{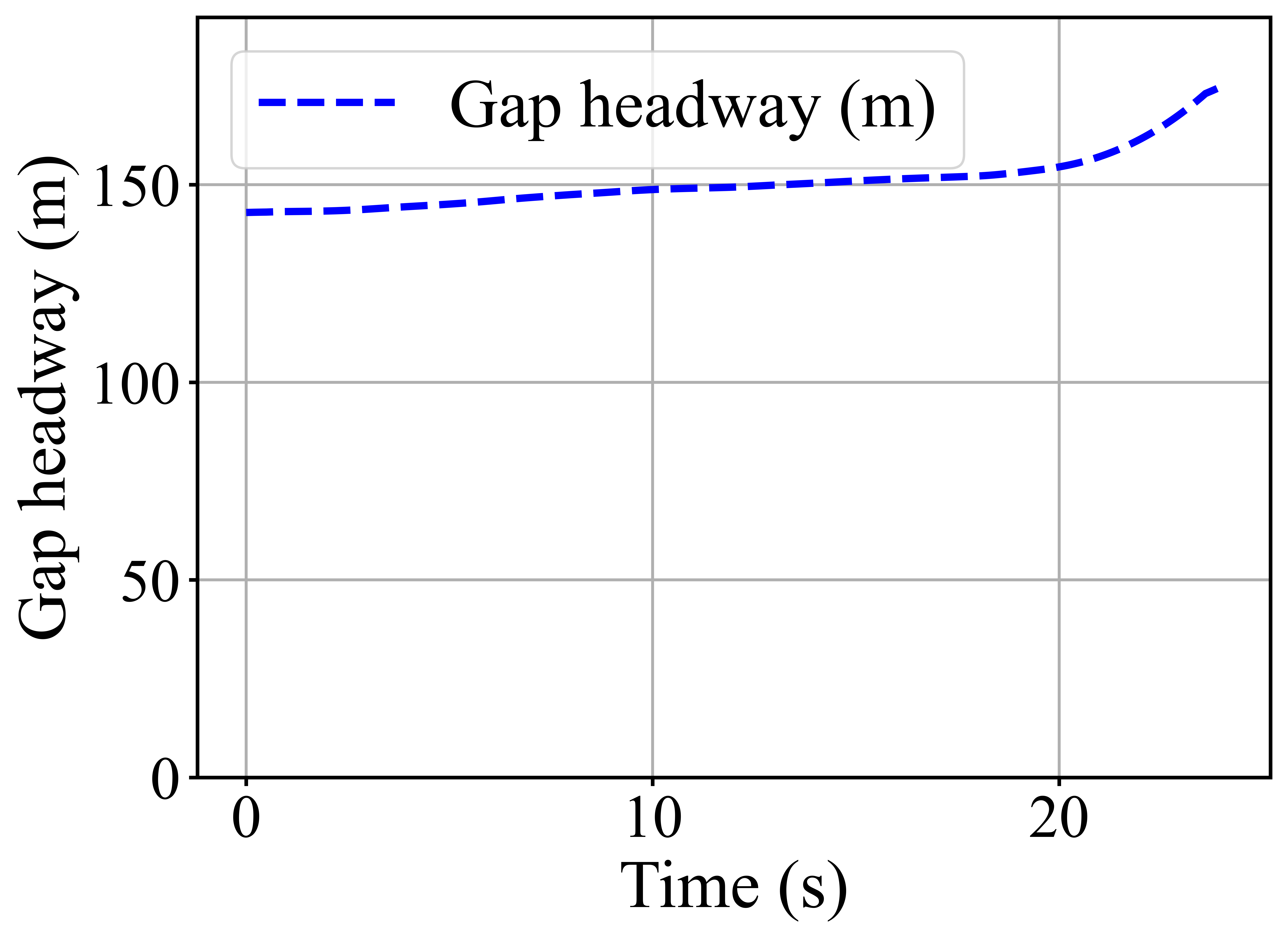}
        \caption{Gap headway 4.}
    \end{subfigure}
    \hfill
    \begin{subfigure}{0.31\textwidth}
    \centering
        \includegraphics[width=\linewidth]{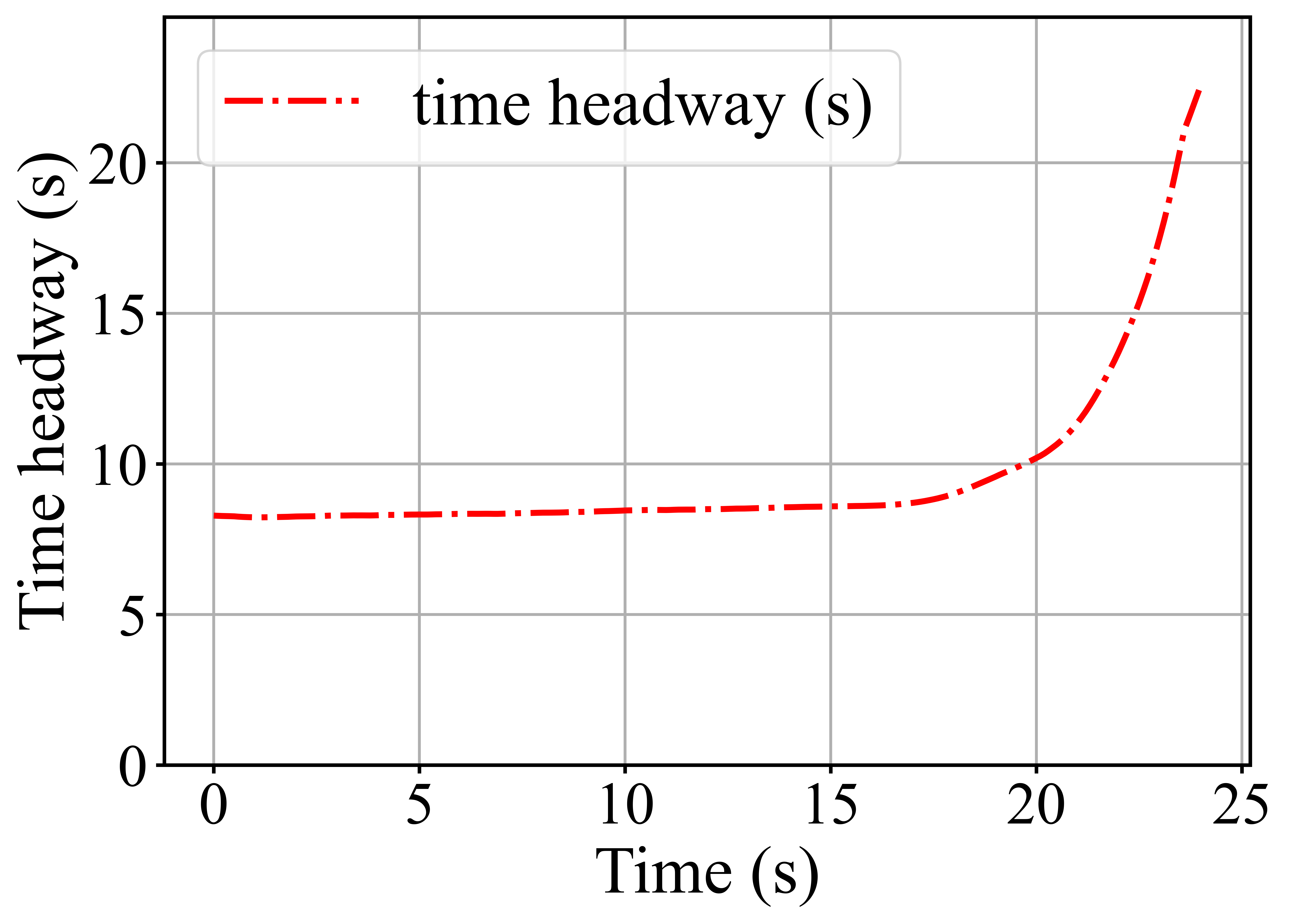}
        \caption{Time headway 4.}
    \end{subfigure}
    \hfill
    \begin{subfigure}{0.31\textwidth}
    \centering
        \includegraphics[width=\linewidth]{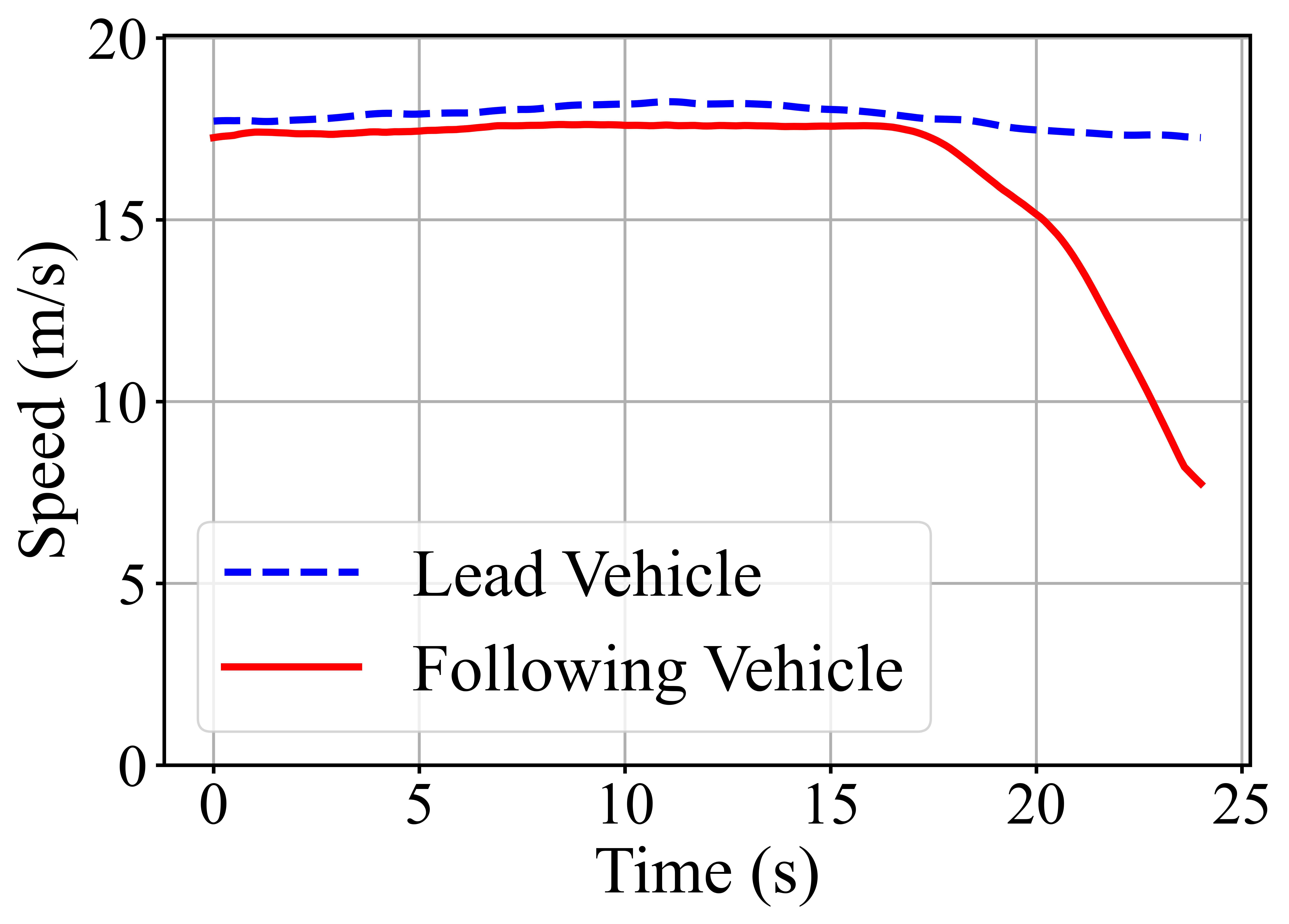}
        \caption{Speed 4.}
    \end{subfigure}
    
    \caption{Car-following threshold determination.}
    \label{fig: Car-following threshold determination}
\end{figure}

\par Figure \ref{fig: Car-following threshold determination} shows the TLSSC-V’s behavior after activation at these four distances. The figure includes the observed gap headway (excluding vehicle lengths), time headway, and speed for each activation distance. In the first two trials (40 m and 60 m), TLSSC-V successfully detected the lead vehicle and followed it through the intersection without decelerating. In contrast, when activated at 90 m and 150 m, TLSSC-V did not detect the lead vehicle, slowed down, and came to a stop at the stop line.

\par From these observations, we conclude that TLSSC-V’s effective detection range for lead vehicles is approximately 90 meters. This threshold can serve as a practical dividing point between car-following and stopping behaviors. While this 90-meter value is not exact due to the simplicity of the experimental setup, it provides a useful estimate of TLSSC-V’s lead vehicle detection limit and can inform future simulation studies of TLSSC behavior.

\section{Conclusion}
\par This work addresses a gap in AV behavior research by empirically characterizing how an advanced ADAS, i.e. the Tesla TLSSC, interacts with TCDs. Through field experiments we built a synchronized trajectory and video dataset, defined a clear behavior taxonomy (stopping, accelerating, and car following), and calibrated the FVDM to each mode, revealing distinct dynamics (e.g., strong responsiveness in stopping versus conservative adjustment in acceleration and smoother, tighter-headway behavior when following through intersections) of the TLSSC-V behaviors. A central finding is the empirical “car-following threshold” of roughly 90 meters: when a lead vehicle is farther than this, TLSSC-V defaults from car-following to a permission-based stopping strategy at green lights. 

\par The dataset and models  developed in this work lay groundwork for broader ADAS–TCD interaction studies, integration into traffic simulations. In future work, we will use the calibrated models in simulation to study the impacts on the traffic system of introducing ADAS-equipped vehicles that interact with TCDs.

\section{Acknowledgements}
This work was supported by the Federal Highway Administration (FHWA) under the Broad Agency Announcement (BAA) Award Number (Grant No. 693JJ324C000003). We gratefully acknowledge the support provided by the FHWA.

\bibliographystyle{cas-model2-names}

\bibliography{cas-refs}

\end{document}